\documentclass{article}

\usepackage[preprint]{corl_2026} 

\usepackage{graphics}
\usepackage{epsfig}
\usepackage{mathptmx}
\usepackage{amsmath, amssymb}

\usepackage{amsthm}
\usepackage{bm}
\usepackage{times}
\usepackage{booktabs}
\usepackage{multirow}
\usepackage{xspace}
\usepackage{algorithm}
\usepackage{enumitem}
\usepackage{subcaption}
\usepackage{algpseudocode}
\usepackage{float}      
\usepackage{graphicx}
\usepackage{fontawesome5}
\usepackage{pifont}
\usepackage{wrapfig}

\DeclareMathOperator*{\argmin}{argmin}

\makeatletter
\newcommand\blfootnote[1]{%
  \begingroup
  \renewcommand\thefootnote{}\footnote{#1}%
  \addtocounter{footnote}{-1}%
  \endgroup
}
\makeatother

\newcommand{\ssf}{\texttt{SafeStreamingFlow}\xspace}

\newtheorem{problem}{Problem}

\newtheorem{theorem}{Theorem}

\newtheorem{definition}{Definition}

\title{Safe Streaming Flow Planning by Aligning Sampling Dynamics with Execution Dynamics}

\author{
  Seunghwan~Jang$^{*\dagger}$\\
  Nanyang Technological University\\
  Singapore\\
  \texttt{seunghwa001@e.ntu.edu.sg} \\
  \And
  Jeongyong~Yang$^{\dagger}$ \\
  University of Washington \\
  Seattle, WA, USA \\
  \texttt{seiryu@uw.edu} \\
  \AND
  Siddharth~Ancha \\
  University of California, Berkeley \\
  Berkeley, CA, USA \\
  \texttt{sancha@berkeley.edu} \\
  \And
  SooJean~Han$^{*}$ \\
  KAIST \\
  Daejeon, Republic of Korea \\
  \texttt{soojean@kaist.ac.kr} \\
}

\begin{document}
\maketitle
\blfootnote{$^{*}$Corresponding authors.}
\blfootnote{$^{\dagger}$Work done at KAIST.}

\begin{abstract}
Generative planners based on diffusion$/$flow matching can learn to synthesize long-horizon trajectories from demonstrations. However, real-world deployment requires (i) enforcing safety constraints during execution and (ii) tight online replanning at fast execution rates.
Prior safe diffusion$/$flow planners generate the agent's full trajectory \textit{at once}, while repeatedly perturbing intermediate states to satisfy safety constraints. This approach is not only computationally intensive, but also introduces distribution shift since the learned sampling dynamics is distinct from the system's execution dynamics.
We propose \ssf, a goal-conditioned planner that aligns flow sampling dynamics with execution dynamics by sequentially integrating a learned state vector field with hierarchical state prediction.
Importantly, we need to enforce safety constraints \emph{only for the executed step} via high order control barrier functions.
Across navigation, racing, and locomotion benchmarks, \ssf reduces planning latency and improves safety compared to existing methods, while maintaining competitive goal-reaching success.
Code is available on our project page: \url{https://jang-seunghwan.github.io/SafeStreamingFlowPlanning/}.
\end{abstract}



\begin{figure*}[h]
\centering
\includegraphics[width=\textwidth]{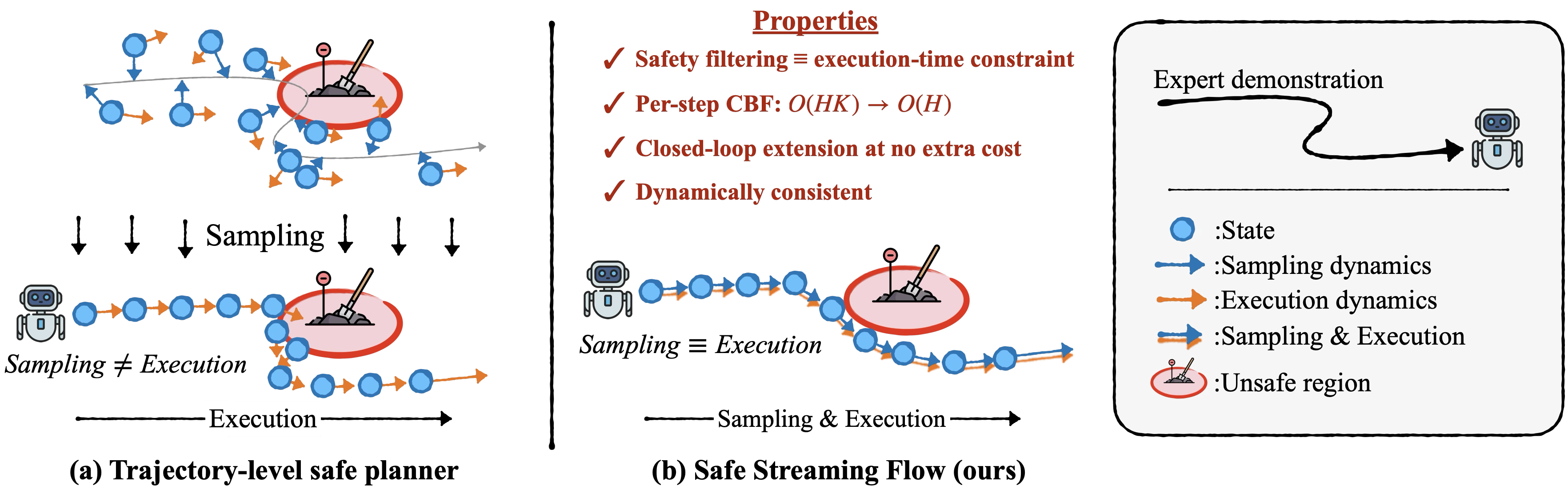}
\caption{\textbf{\ssf{} aligns flow sampling dynamics with execution dynamics.}
\textbf{(a)} Trajectory-level safe planners denoise an entire trajectory along
a sampling process distinct from the system's execution dynamics (blue and orange arrows misaligned).
Because these intermediate sampling states are not physically executed, enforcing safety during sampling disrupts dynamical consistency and requires $\mathcal O(HK)$ optimization solves.
\textbf{(b)} \ssf integrates a learned state vector field in physical time, so each sampling step coincides with an executed step (blue and orange aligned). A safety constraint imposed on the sampling dynamics is therefore a constraint on the executed state under the system model, yielding per-step safety enforcement with $\mathcal{O}(H)$ optimization solves (versus $\mathcal{O}(HK)$), closed-loop execution without architectural or computational change, and dynamically consistent plans whose executed trajectory matches the generated one by construction.}
\label{fig:teaser}
\end{figure*}

\section{Introduction}

Generative models~\cite{ho2020denoising, lipman2022flow} that have driven rapid progress in image and language generation are increasingly being adapted to robotic tasks, particularly diffusion~\cite{janner2022planning,fahnestock2024learning,jiang2025anomalies} and flow matching~\cite{yang2026safeflowmatcher}.
Deploying such models for robot planning, however, requires more than producing a plausible trajectory.
%
Most diffusion/flow planners typically generate an entire horizon as a single object in trajectory space and refine it through $K$ denoising/sampling steps, where each step updates all $H$ states in the trajectory~\cite{janner2022planning}.
In a closed-loop setting, this full-horizon procedure must be repeated at every control cycle to incorporate state feedback and new observations, generating $H\times K$ intermediate states per cycle~\cite{jiang2025streaming,hoeg2025fast}.

Another key challenge is enforcing safety constraints imposed at inference time, possibly never seen during training, in a manner that is both reliable and compatible with closed-loop execution. Many generative methods promote safety through guidance~\cite{dhariwal2021diffusion,ho2021classifierfree}, injecting heuristic costs, penalties, or constraint-related scores into the generation process to steer samples away from unsafe regions; such guidance either requires the constraint to be learned in advance or imposes it only softly, without formal guarantees~\cite{pmlr-v305-jung25d,pmlr-v305-ma25a}.
An especially promising approach to \emph{certified safety} is to incorporate control barrier functions (CBFs)~\cite{ames2016control,ames2019control} into the sampling procedure by solving quadratic programs (QPs) at inference time; the constraints then hold whenever each QP is feasible.
However, because intermediate sampling states are not physically executed, enforcing safety during sampling can disrupt dynamical consistency of the generated trajectories and induce \emph{local-traps}, discontinuous trajectories that remain safe but infeasible to execute~\cite{yang2026safeflowmatcher, xiao2025safediffuser}.

In addition, safety enforcement further increases the computational burden because CBF-based corrections are applied across both the horizon length $H$ and sampling iterations $K$, resulting in $(H\times K)$ QP solves to generate a single trajectory.
\emph{Streaming-based} formulations~\cite{hoeg2025fast, jiang2025streaming} have been introduced for real-time execution by aligning sampling time with execution time.
However, existing streaming approaches are developed primarily as \emph{policies} for action imitation, leaving both the system's execution dynamics and start--goal-conditioned \emph{planning} unaddressed.

We propose \ssf, a safe streaming planner that aligns the generative process with execution (Figure~\ref{fig:teaser}).
Instead of sampling at the full trajectory level, \ssf integrates a learned state vector field in physical time and produces a one-step-ahead reference.
In this one-step prediction, however, the learned field tends to become a local inertial extrapolator rather than a goal-conditioned planner; we suppress this with a hierarchical state predictor that sequentially updates position then velocity.
The one-step-ahead reference is then tracked by a PD controller, and the resulting nominal control input is filtered through high order control barrier functions that minimally modify the executed input to satisfy safety constraints that were never seen during training.

Table~\ref{tab:paradigm_comparison} positions \ssf along four properties that classical and generative planners typically trade off: a cost-function-free formulation, multimodal trajectory generation, execution-time safety, and real-time streaming.
Classical planners (RRT*, A*) satisfy safety constraints by construction but require engineered costs and produce a single non-streaming plan; generative planners (Diffuser, SafeDiffuser) gain cost-function-free multimodality but offer only partial, non-streaming safety. \ssf is the only paradigm attaining all four.

\begin{table}[h]
\centering
\caption{Conceptual positioning of \ssf against existing planning paradigms.}
\vspace{4px}
\label{tab:paradigm_comparison}
\small
\setlength{\tabcolsep}{4pt}
\resizebox{0.8\textwidth}{!}{%
\begin{tabular}{l ccccc}
\toprule
\textbf{Paradigm} 
& \textbf{Cost-fn.\ free}
& \textbf{Multimodal}
& \textbf{Exec.\ safety}
& \textbf{Real-time} \\
\midrule
Classical (RRT*, A*)                & \ding{55}    & \ding{55} & \checkmark   & $\triangle$ \\
Generative (Diffuser, SafeDiffuser) & \checkmark   & \checkmark  & $\triangle$  & \ding{55} \\
\textbf{\ssf (Ours)}              & \checkmark   & \checkmark & \checkmark   & \checkmark \\
\bottomrule
\end{tabular}%
}
\end{table}


\section{Related Work}
Generative models such as diffusion~\cite{ho2020denoising} and flow matching~\cite{lipman2022flow} have recently been adopted for robot policy learning and planning by modeling multi-modal state/action distributions from expert demonstrations~\cite{janner2022planning,chi2024diffusionpolicy, carvalho2023motion}.
Diffusion-based planners synthesize long-horizon plans through iterative denoising in trajectory space, while flow matching reduces the number of sampling steps via ODE-based generation~\cite{ye2024efficient}.
More recently, streaming formulations have been proposed to better align generation with closed-loop control by producing incremental updates~\cite{prasad2024consistency, jiang2025streaming, hoeg2025fast}; however, these methods are primarily formulated as action policies and do not directly address the system's execution dynamics or finite-horizon, start--goal-conditioned planning.

Safety in generative planning is commonly incorporated either through guidance~\cite{pmlr-v305-jung25d,pmlr-v305-ma25a,bodmer2026groundinggenerativepoliciesphysics}, such as classifier/classifier-free guidance~\cite{dhariwal2021diffusion, ho2021classifierfree} and CBF-shaped objectives~\cite{mizuta2024cobl}, or through CBF-based safety filters embedded in the sampling pipeline~\cite{yang2026safeflowmatcher,xiao2025safediffuser}.
While guidance-based methods generally lack formal guarantees, safety filters provide them under their modeling assumptions: Hamilton--Jacobi reachability~\cite{bansal2017hamilton} scales exponentially with the state dimension and model predictive safety filters~\cite{wabersich2021predictive} solve a horizon-length problem at every step, whereas CBF filters~\cite{ames2016control,ames2019control} need a single optimization per control cycle.
However, trajectory-level CBF correction can introduce computational overhead, local traps, and dynamical inconsistency.
In contrast, \ssf extends streaming flow to start--goal-conditioned state space planning and enforces safety constraints only at the executed step, aligning generation, execution, and safety filtering in physical time.


\section{Problem Formulation}\label{sec:problem}
\subsection{Setup}
We consider a control-affine dynamical system of the form
\begin{equation}\label{eq:dynamics}
    \dot{\mathbf{x}}_t = f(\mathbf{x}_t)+g(\mathbf{x}_t)\mathbf{u}_t=
    \begin{bmatrix}
    \mathbf{v}_t \\ \tilde f(\mathbf{p}_t,\mathbf{v}_t)
    \end{bmatrix} +
    \begin{bmatrix}
    \mathbf{0} \\ \tilde g(\mathbf{p}_t,\mathbf{v}_t)
    \end{bmatrix}
    \mathbf{u}_t,
\end{equation}
where $\mathbf{x}_t = (\mathbf{p}_t, \mathbf{v}_t) \in \mathcal{X} \subset \mathbb{R}^{2n}$ denotes the system state ($\mathcal{X}$ is the feasible state space), $\mathbf{p}_t \in \mathbb{R}^n$ and $\mathbf{v}_t \in \mathbb{R}^n$ are position and velocity, respectively, and $\mathbf{u}_t \in \mathcal{U} \subset \mathbb{R}^n$ is the control input ($\mathcal{U}$ is the admissible control space). The functions $f:\mathcal{X}\to\mathbb{R}^{2n}$ and $g:\mathcal{X}\to\mathbb{R}^{2n\times n}$ denote the full state drift and control vector fields, respectively. We use $\tilde f:\mathcal{X}\to\mathbb{R}^n$ and $\tilde g:\mathcal{X}\to\mathbb{R}^{n\times n}$ to denote their acceleration-level components.
All functions are assumed to be locally Lipschitz continuous.
This formulation captures a broad class of second-order systems, including point-mass and mass--spring--damper dynamics, and is widely used as a tractable approximation of more complex robotic and mechanical systems, where the control input acts directly on the acceleration dynamics (i.e., $\tilde{g}(\mathbf{p},\mathbf{v})$ has full column rank).

Let $T \in \mathbb{R}_{\geq 0}$ denote the planning horizon time. A trajectory is defined as $\bm{\tau} : [0,T] \to \mathcal{X}$, i.e., $\bm{\tau}=\{\mathbf{x}_t : t \in [0,T]\}$.
We are given a dataset of $N$ expert demonstrations $\mathcal{D} = \{\bm{\tau}^{(j)}\}_{j=1}^N$, where each demonstration starts at $\mathbf{x}_0^{(j)}$ and terminates at a goal state $\mathbf{x}_T^{(j)}$.
In addition, we define the sampling time $\Delta t = T/H$ and the discrete time $k=0,\ldots,H$, corresponding to time $k\Delta t$ ($\mathcal{T}\triangleq\{0,1,\ldots,H-1\}$).
Our safety filter and its guarantee are stated for the Euler discretization of \eqref{eq:dynamics}:
\begin{equation}\label{eq:dt_dynamics}
    \mathbf{x}_{k+1} = \mathbf{x}_k + \Delta t \big(f(\mathbf{x}_k) + g(\mathbf{x}_k)\mathbf{u}_k \big).
\end{equation}

Safety constraints are specified by twice continuously differentiable barrier functions $\{h_i:\mathcal{X}\to\mathbb{R}\}_{i\in\mathcal{I}}$, where $\mathcal{I}$ is a finite index set of constraints.
The safe set is defined as the intersection
\begin{equation}\label{eq:safe_set}
    \mathcal{C} \triangleq \bigcap_{i\in\mathcal{I}} \big\{\mathbf{x} \in \mathcal{X} : h_i(\mathbf{x}) \geq 0\big\}.
\end{equation}
Here, we focus on constraints of relative degree two with respect to the system~\eqref{eq:dynamics}, i.e., $L_g h(x) = 0$ and $L_g L_f h(x) \neq 0$ on $\mathcal{C}$.
Under \eqref{eq:dt_dynamics}, $\mathbf{p}_{k+1}=\mathbf{p}_k+\Delta t\,\mathbf{v}_k$ does not involve $\mathbf{u}_k$, so the input first affects $h_i$ at step $k+2$ and a first-order condition on $h_i$ alone cannot be enforced through $\mathbf{u}_k$.
\begin{problem}[Safe planning from demonstrations]\label{main_problem}
    Given the system dynamics \eqref{eq:dt_dynamics}, demonstration dataset $\mathcal{D}$, start state $\mathbf{x}_s=(\mathbf{p}_s,\mathbf{v}_s)$, goal state $\mathbf{x}_g=(\mathbf{p}_g,\mathbf{v}_g)$, planning horizon $T$, and a safe set $\mathcal{C}$ defined in \eqref{eq:safe_set}, design a planner that generates a state trajectory $\{ \mathbf{x}_0, \mathbf{x}_1, \ldots, \mathbf{x}_H \}$ such that $\mathbf{x}_0=\mathbf{x}_s$, $\|\mathbf{x}_H-\mathbf{x}_g\|_2\le\varepsilon_g$ for some goal tolerance $\varepsilon_g>0$, and $\mathbf{x}_k\in \mathcal{C}$ for all $k = 0, \ldots, H$, and the state trajectory satisfies the system dynamics~\eqref{eq:dt_dynamics}.
\end{problem}

\subsection{Preliminaries: Discrete-Time High Order Control Barrier Functions}
We review discrete-time high order control barrier functions (HOCBFs)~\cite{xiong2023discretehocbf}, focusing on relative degree two constraints for the discretized dynamics \eqref{eq:dt_dynamics}.
For each $i\in\mathcal{I}$, define the discrete-time barrier sequence
\begin{subequations}
\begin{align}
    \psi_{i,0}(\mathbf{x}_k) &\triangleq h_i(\mathbf{x}_k), \label{eq:dt_psi0} \\
    \psi_{i,1}(\mathbf{x}_k) &\triangleq \psi_{i,0}(\mathbf{x}_{k+1}) - \psi_{i,0}(\mathbf{x}_k) + \alpha_{i,1}\big(\psi_{i,0}(\mathbf{x}_k)\big), \label{eq:dt_psi1} \\
    \psi_{i,2}(\mathbf{x}_k,\mathbf{u}_k) &\triangleq \psi_{i,1}(\mathbf{x}_{k+1}) - \psi_{i,1}(\mathbf{x}_k) + \alpha_{i,2}\big(\psi_{i,1}(\mathbf{x}_k)\big), \label{eq:dt_psi2}
\end{align}
\end{subequations}
where $\mathbf{x}_{k+1}$ follows~\eqref{eq:dt_dynamics}.
By the relative-degree-two structure, $\psi_{i,1}$ is a function of the state alone, so the sets below are well defined, and $\mathbf{u}_k$ enters only $\psi_{i,2}$, through $\mathbf{p}_{k+2}$ (explicit form in Appendix~\ref{appendix:qcqp}). $\alpha_{i,1}$ and $\alpha_{i,2}$ are class-$\mathcal{K}$ functions satisfying $\alpha(s)<s$ for $s>0$.

We also define the associated sets
\begin{equation}
    \mathcal{C}_{i,0} \triangleq \{\mathbf{x}_k : \psi_{i,0}(\mathbf{x}_k) \geq 0\}, \quad \mathcal{C}_{i,1} \triangleq \{\mathbf{x}_k : \psi_{i,1}(\mathbf{x}_k) \geq 0\},
\end{equation}
and their augmented set $\mathcal{C}_i \triangleq \mathcal{C}_{i,0} \cap \mathcal{C}_{i,1}$.

\begin{definition}[Discrete-Time HOCBFs with Relative Degree Two]\label{def:hocbf}
A twice continuously differentiable function $h_i$ is a discrete-time HOCBF of relative degree two for system~\eqref{eq:dt_dynamics} if there exist class-$\mathcal{K}$ functions $\alpha_{i,1}$ and $\alpha_{i,2}$ such that for every $\mathbf{x}_k\in\mathcal{C}_i$ there exists $\mathbf{u}_k\in\mathcal{U}$ with
\begin{equation}\label{eq:dt_hocbf_condition}
    \psi_{i,2}(\mathbf{x}_k,\mathbf{u}_k) \geq 0.
\end{equation}
\end{definition}
\begin{theorem}\label{thm:invariance}
    Consider the discrete-time system \eqref{eq:dt_dynamics}.
    If $\mathbf{x}_0 \in \bigcap_{i\in\mathcal{I}} \mathcal{C}_{i}$ and the applied input $\mathbf{u}_k \in \mathcal{U}$ satisfies \eqref{eq:dt_hocbf_condition} for all $i\in\mathcal{I}$ at every $k\ge 0$, then $\bigcap_{i\in\mathcal{I}} \mathcal{C}_{i}$ is forward invariant along the resulting trajectory\cite{xiong2023discretehocbf}, i.e., $\mathbf{x}_k\in\bigcap_{i\in\mathcal{I}}\mathcal{C}_i\subseteq\mathcal{C}$ for all $k\ge 0$.
\end{theorem}
%

In contrast to continuous-time CBF constraints~\cite{ames2016control}, the discrete-time HOCBF constraint~\eqref{eq:dt_hocbf_condition} is not necessarily affine in the control input. We can find a safe control input by solving the following optimization problem that minimally modifies a nominal input $\mathbf{u}^{\mathrm{nom}}$:
\begin{equation}\label{eq:hocbf_qcqp_prelim}
    \mathbf{u}^{\mathrm{safe}} = \argmin_{\mathbf{u}\in\mathcal{U}} \ \|\mathbf{u}-\mathbf{u}^{\mathrm{nom}}\|_2^2 \quad \text{s.t. } \eqref{eq:dt_hocbf_condition} \text{ for all } i\in\mathcal{I}.
\end{equation}
When $h_i$ is quadratic in $\mathbf{x}$, the optimization problem can be formulated as a quadratically constrained quadratic program (QCQP).
Theorem \ref{thm:invariance} (proof in Appendix \ref{app:proof_thm1}) requires \eqref{eq:hocbf_qcqp_prelim} to be feasible at every step, and we do not establish this recursive feasibility for our hand-specified barriers.


\section{Safe Streaming Flow Planning}\label{sec:ssfp}

\begin{figure}[t]
\centering
\includegraphics[width=\linewidth]{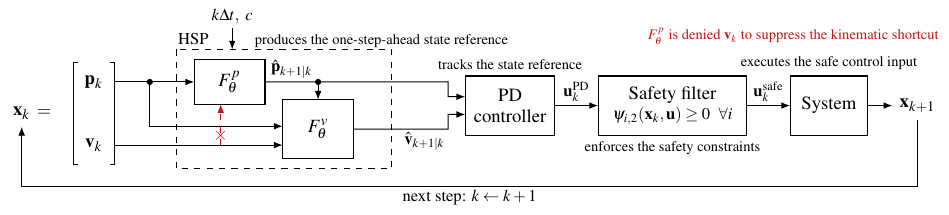}
\vspace{-20px}
\caption{One step of \ssf, following Eqs.~\eqref{eq:HSP_pos}--\eqref{eq:hocbf_qcqp}.}
\vspace{-20px}
\label{fig:ssf_step}
\end{figure}

\ssf has two components (Figure \ref{fig:ssf_step}). The first is a streaming generator of state trajectories from a learned start--goal-conditioned vector field: Section~\ref{subsec:sfp} extends streaming flow---developed in prior work as an action policy~\cite{jiang2025streaming}---to start--goal-conditioned state space planning, and Section~\ref{subsec:hsp} introduces a hierarchical architecture for $F_\theta$ that suppresses a kinematic shortcut emerging during training on second-order systems. The second is a safety filter applied to the executed control input, which Section~\ref{subsec:cbf} formulates using a discrete-time HOCBF filter.

\subsection{Streaming Flow Planning}\label{subsec:sfp}
We formulate planning as streaming generation by integrating a learned start--goal-conditioned vector field in physical time
\begin{equation}\label{eq:ssf_field}
    \dot{\mathbf{x}}_t = F_\theta\bigl(\mathbf{x}_t, t \mid c\bigr), \quad t\in[0,T],
\end{equation}
where $F_\theta$ is a neural vector field and $c\triangleq(\mathbf{x}_s, \mathbf{x}_g)$. 
Nominal trajectories are generated by integrating \eqref{eq:ssf_field} forward over the sampling time $t\in[0,T]$\footnote{Most flow matching formulations use a normalized time variable $s\in[0,1]$. Instead, we treat flow sampling time as identical to physical time $t\in[0,T]$. This is an equivalent re-parameterization corresponding to $s=t/T$, which yields $\frac{d\bm{\tau}}{dt} = \frac{1}{T}\frac{d\bm{\tau}}{ds}$.}.
Unlike trajectory-level diffusion/flow planners that generate an entire horizon through an inner sampling process, \ssf identifies generation (sampling) time with planning time. In discrete time, a one-step forward state is generated by Euler integration
\begin{equation}\label{eq:discrete_integration}
    \mathbf{x}_{k+1}
    = \mathbf{x}_k + \Delta t\,F_\theta(\mathbf{x}_k,k\Delta t\mid c), \qquad k\in\mathcal{T}.
\end{equation}
For multimodal plans, the initial position can be sampled near $\mathbf{p}_s$ (Appendix \ref{app:multimodality}).

We train the vector field by constructing contracting conditional flows \cite{jiang2025streaming} around expert demonstrations.
Let $A_{\Delta t} \triangleq [I_n, \Delta t^{-1} I_n ]^{\top} \in \mathbb{R}^{2n \times n}$ denote the kinematic lift mapping a position-space perturbation to a state space perturbation.
For a demonstration $\bm{\tau}(t)=(\mathbf{p}_{\bm{\tau}}(t),\mathbf{v}_{\bm{\tau}}(t))$, we define the target field
\begin{equation}\label{eq:tau_target_field}
    F_{\bm{\tau}}(\mathbf{x},t) \triangleq \dot{\bm{\tau}}(t) - \lambda A_{\Delta t} \bigl(\mathbf{p}-\mathbf{p}_{\bm{\tau}}(t)\bigr), \qquad \mathbf{x}=(\mathbf{p},\mathbf{v}),
\end{equation}
where $\lambda \ge 0$ is a contracting gain.
This construction embodies two design choices. 
\textbf{(i)} To learn a vector field consistent with the kinematic dynamics $\mathbf{v}=\dot{\mathbf{p}}$, perturbations are confined to the kinematically coupled subspace via $A_{\Delta t}$, which maps a position perturbation $\delta\mathbf{p}$ to the induced velocity perturbation $\delta\mathbf{v}=\delta\mathbf{p}/\Delta t$.
\textbf{(ii)} The resulting model is a \emph{state-reference planner}, explicitly conditioned on $(\mathbf{x}_s,\mathbf{x}_g)$ and horizon $T$ and emitting a state trajectory tracked by a low-level controller---not a reactive policy over abstract action setpoints as in~\cite{jiang2025streaming}.

Since the planner must end at $\mathbf{x}_g$ (Problem \ref{main_problem}); we therefore show that the minimizer of the training objective below matches the per-timestep marginals of the conditional paths (Theorem \ref{thm:marginal_matching}), which yields terminal goal-reaching at $t=T$.
For $\boldsymbol{\xi}\sim\mathcal{N}(0,I_{n})$, define the trajectory-conditioned state sample
\begin{equation}\label{eq:traj_cond_state_sample}
    \mathbf{x}_t^{\bm{\tau}} = [\mathbf{p}_{\mathbf{x}_t^{\bm{\tau}}}, \mathbf{v}_{\mathbf{x}_t^{\bm{\tau}}}]^\top = \bm{\tau}(t) + \sigma_t A_{\Delta t}\boldsymbol{\xi}, \qquad \sigma_t=\sigma_0 e^{-\lambda t},
\end{equation}
and denote its induced distribution by $\mu_t^{\bm{\tau}}$. Then, we train $F_\theta$ using conditional flow matching loss:
\begin{equation}\label{eq:cfm_loss}
    \mathcal{L}(\theta) =
    \mathbb{E}_{\bm{\tau}\sim\mathcal{D},\, t\sim\mathrm{Unif}(0,T),\, \boldsymbol{\xi}\sim\mathcal{N}(0,I_n)} \left[ \left\| F_\theta(\mathbf{x}_t^{\bm{\tau}},t \mid \bm{\tau}(0),\bm{\tau}(T)) - F_{\bm{\tau}}(\mathbf{x}_t^{\bm{\tau}},t) \right\|_2^2 \right].
\end{equation}
Thus, $F_\theta$ learns a start--goal-conditioned vector field by regressing onto trajectory-conditioned fields sampled along the prescribed conditional paths.

\begin{theorem}\label{thm:marginal_matching}
    Let $p_{\mathcal{D}}(\cdot\mid c)$ denote the $c$-conditional distribution induced by the dataset $\mathcal{D}$, and let $F^*$ be the minimizer of the conditional flow matching loss \eqref{eq:cfm_loss}. Then, for every $t \in [0,T]$ and a fixed start--goal condition $c=(\mathbf{x}_s, \mathbf{x}_g)$, the flow induced by $F^*$ transports $\mu_0^c$ to the target per-timestep marginal distribution $\mu_t^c = \mathbb{E}_{\bm{\tau}\sim p_{\mathcal{D}}(\cdot \mid c)} [\mu_t^{\bm{\tau}}]$.
\end{theorem}
The detailed proof of Theorem~\ref{thm:marginal_matching} is provided in Appendix~\ref{appendix:proof_marginal}.


As an immediate consequence, suppose that every demonstration in $p_{\mathcal{D}}(\cdot\mid c)$ satisfies $\bm{\tau}(T)=\mathbf{x}_g$.
Substituting this into~\eqref{eq:traj_cond_state_sample} at $t=T$ yields 
$\mathbf{x}_T^{\bm{\tau}}=\mathbf{x}_g+\sigma_T A_{\Delta t}\boldsymbol{\xi}$, 
independent of $\bm{\tau}$. The matched terminal marginal is therefore the 
pushforward of $\mathcal{N}(0,I_n)$ under this affine map,
\begin{equation}\label{eq:goal_concentration}
\mu_T^c \;=\; \bigl(\mathbf{x}_g + \sigma_T A_{\Delta t}\,\cdot\,\bigr)_{\#}\,\mathcal{N}(0,I_n),
\end{equation}
whose position component is $\Pi_p\mathbf{x}_T^{\bm{\tau}}\sim\mathcal{N}(\mathbf{p}_g,\sigma_T^2 I_n)$ 
(since $\Pi_p A_{\Delta t}=I_n$), giving position error 
$\mathbb{E}\|\Pi_p\mathbf{x}_T^{\bm{\tau}}-\mathbf{p}_g\|_2^2 = n\sigma_T^2$.
Hence, step-wise marginal matching 
(Theorem~\ref{thm:marginal_matching}) implies terminal goal-reaching at scale 
$O(\sigma_T)$ under goal-curated demonstrations, although it leaves the joint law of the path undetermined.

\subsection{Hierarchical State Predictor (HSP)}\label{subsec:hsp}
We now specify the parametric form of $F_\theta$ from~\eqref{eq:ssf_field}. 
Exposing the full state $(\mathbf{p}_t,\mathbf{v}_t)$ to a single network admits a \emph{kinematic shortcut}: for locally bounded acceleration,
\begin{equation}\label{eq:hsp_first_order_position}
    \mathbf{p}_{t+\Delta t} = \mathbf{p}_t + \Delta t\,\mathbf{v}_t + \mathcal{O}(\Delta t^2),
\end{equation}
so a predictor can drive training loss down by carrying over $\mathbf{v}_t$ rather than learning goal-conditioned progress from $c=(\mathbf{x}_s,\mathbf{x}_g)$ (ablated in Appendices~\ref{app:hsp_ablation} and~\ref{app:hsp_justification}).
The resulting field becomes a local inertial extrapolator rather than a goal-conditioned planner: predicting where the trajectory inertially \emph{will be} one-step ahead, not where it \emph{should be} under the start--goal plan.

To suppress this, we factorize $F_\theta=(F^p_\theta,F^v_\theta)$ and sequentially update position then velocity (Figure \ref{fig:ssf_step}):
\begin{align}
    \hat{\mathbf{p}}_{k+1\mid k} &= \mathbf{p}_k + \Delta t\,F^p_\theta(\mathbf{p}_k,\, k\Delta t \mid c), \label{eq:HSP_pos}\\
    \hat{\mathbf{v}}_{k+1\mid k} &= \mathbf{v}_k + \Delta t\,F^v_\theta(\mathbf{p}_k,\, \mathbf{v}_k,\, \hat{\mathbf{p}}_{k+1\mid k},\, k\Delta t \mid c). \label{eq:HSP_vel}
\end{align}
Both branches are trained by~\eqref{eq:cfm_loss} applied componentwise to $F_{\bm\tau}$~\eqref{eq:tau_target_field}: $F^p_\theta$ regresses onto its position component, $F^v_\theta$ onto its velocity component.

HSP is thus a structured parameterization of the same $F_\theta$ as in Section~\ref{subsec:sfp}, not a new objective.
The only change is that $F^p_\theta$ is denied $\mathbf{v}_k$, forcing it to infer $\hat{\mathbf{p}}_{k+1\mid k}$ from the geometric planning context $(\mathbf{p}_k, k\Delta t, \mathbf{x}_s, \mathbf{x}_g)$ while $F^v_\theta$ restores local dynamic consistency conditioned on the predicted next position. This separation is what restores goal sensitivity and closed-loop robustness, isolated causally in Appendix~\ref{app:hsp_justification}. We note that~\eqref{eq:HSP_pos}--\eqref{eq:HSP_vel} specify the predictor parameterization, not the physical state transition law; a first-order local update is used since the exact coupled one-step map is unavailable in closed form.

\subsection{Safe Planning via HOCBFs}\label{subsec:cbf}
We couple streaming state generation with a PD controller and safety filter to form a complete planning and execution loop. At each execution cycle, the generative model produces a one-step-ahead state reference $(\hat{\mathbf{p}}_{k+1 \mid k}, \hat{\mathbf{v}}_{k+1 \mid k})$ via \eqref{eq:HSP_pos}--\eqref{eq:HSP_vel}.
This reference is tracked by a PD controller, whose nominal control input is
\begin{equation}\label{eq:pd_control}
    \mathbf{u}_k^{\mathrm{PD}} = K_p (\hat{\mathbf{p}}_{k+1 \mid k}-\mathbf{p}_k) + K_v (\hat{\mathbf{v}}_{k+1 \mid k}-\mathbf{v}_k)
\end{equation}
where $K_p$ and $K_v$ are proportional and derivative gains, respectively.

We compute the safe control input by solving the following optimization problem similar to~\eqref{eq:hocbf_qcqp_prelim}:
\begin{equation}\label{eq:hocbf_qcqp}
    \mathbf{u}_k^{\mathrm{safe}} = \argmin_{\mathbf{u}_k\in\mathcal{U}} \|\mathbf{u}_k-\mathbf{u}_k^{\mathrm{PD}}\|_2^2 \quad \text{s.t. } \eqref{eq:dt_hocbf_condition} \text{ for all } i\in\mathcal{I}.
\end{equation}
This optimization returns a control input that minimally deviates from the PD input while satisfying the
safety constraints. Since \eqref{eq:hocbf_qcqp} is generally a nonlinear programming problem, we consider quadratic barrier functions in our experiments.
We evaluate the relative-degree-two discrete-time HOCBF using the two-step local Euler prediction $\mathbf{p}_{k+2\mid k}(\mathbf{u}_k)$ derived in Appendix~\ref{appendix:qcqp}.

Finally, we propagate the model \eqref{eq:dt_dynamics} by using $\mathbf{u}_k^{\mathrm{safe}}$ to obtain the next state $\mathbf{x}_{k+1}^{\mathrm{safe}}$, naturally maintaining dynamical consistency.
This safe state is then fed back into the vector field for the next streaming update; in closed-loop mode, the measured state is fed back instead.
Since \ssf generates trajectories in a streaming manner, it can be used in both open-loop and closed-loop (online) modes; the full procedure is summarized in Appendix~\ref{appendix:alg}, with the mode selected by $\Delta$.

\section{Experiments}
\label{sec:result}

We compare \ssf against safe generative planners in both nominal and safe variants: guidance-based (classifier guidance, CG~\cite{dhariwal2021diffusion}), diffusion-based (Diffuser/SafeDiffuser~\cite{janner2022planning,xiao2025safediffuser}), and flow-matching-based (FM/FlowMatcher and SafeFM/SafeFlowMatcher~\cite{yang2026safeflowmatcher}). 
We evaluate open- and closed-loop planning on four environments: Maze2D~\cite{fu2020d4rl} (Section~\ref{subsec:maze2d}), F1TENTH~\cite{kelly2020f1tenth} (Section~\ref{subsec:f1tenth}), MuJoCo Hopper~\cite{fu2020d4rl} (Section~\ref{subsec:hopper}), and a ROS2/Gazebo warehouse navigation~\cite{gazebo} (Section~\ref{subsec:gazebo}).

Throughout, $H$ denotes the discrete horizon length (i.e., the number of transitions) and $K$ the number of sampling steps a trajectory-level baseline used to generate a full trajectory.
We note that SafeFlowMatcher uses $K{+}1$ because of its two-phase structure~\cite{yang2026safeflowmatcher}; the specific values of $H$ and $K$ used per task are listed in each results table. In contrast, \ssf emits one state per sampling step, requiring $H$ sequential forward passes irrespective of $K$.
We report success rate (\emph{Succ.}), the plan-level indicator $\bar h_{\min}{>}0$, the execution-level violation rate (\emph{Viol.}), acceleration 
smoothness (\emph{Sm.}), trap rate (\emph{Trap.}), and timing ($t_{\text{Opt}}$, 
$t_{\text{total}}$; per trajectory, and per action in Hopper).
All metrics are averaged over 100 rollouts; see Appendix~\ref{appendix:experiment}.

\subsection{Maze2D}
\label{subsec:maze2d}

\begin{table*}[h]
\caption{Maze2D results across Umaze, Medium, Large maps.}
\label{tab:maze2d_main}
\centering
\footnotesize
\resizebox{\textwidth}{!}{%
\begin{tabular}{l l ccccc ccccc ccccc}
\toprule
& & \multicolumn{5}{c}{\textbf{Umaze ($H=128$; $K=64$)}} & \multicolumn{5}{c}{\textbf{Medium ($H=256$; $K=256$)}} & \multicolumn{5}{c}{\textbf{Large ($H=384$; $K=256$)}} \\
\cmidrule(lr){3-7} \cmidrule(lr){8-12} \cmidrule(lr){13-17}
\textbf{Category} & \textbf{Method}
  & Succ. & $\bar h_{\min}{>}0$ & Viol. & Sm. & Trap
  & Succ. & $\bar h_{\min}{>}0$ & Viol. & Sm. & Trap
  & Succ. & $\bar h_{\min}{>}0$ & Viol. & Sm. & Trap \\
\midrule
  \multirow{2}{*}{Classical}
   & RRT*~\cite{karaman2011rrtstar}
       & 91.0 & \checkmark & 5.0 & $1.980{\scriptstyle\pm2.582}$ & -
       & 94.0 & \checkmark & 1.0 & $0.629{\scriptstyle\pm0.678}$ & -
       & 86.0 & \checkmark & 8.0 & $0.592{\scriptstyle\pm0.548}$ & - \\
   & A*~\cite{hart1968astar}
       & 93.0 & \checkmark & 7.0 & $2.613{\scriptstyle\pm3.404}$ & -
       & 85.0 & \checkmark & 14.0 & $1.377{\scriptstyle\pm1.143}$ & -
       & 85.0 & \checkmark & 6.0 & $0.689{\scriptstyle\pm0.628}$ & - \\
\midrule
  \multirow{3}{*}{Diffusion}
   & Diffuser~\cite{janner2022planning} & $94.0^\dagger$ & - & 38.0 & $57.560{\scriptstyle\pm19.586}$ & - & $94.0^\dagger$ & - & 53.0 & $59.995{\scriptstyle\pm11.655}$ & - & $74.0^\dagger$ & - & 65.0 & $59.939{\scriptstyle\pm8.099}$ & - \\
   & Diffuser + CG~\cite{dhariwal2021diffusion} & 36.0 & \checkmark & 51.0 & $302.012{\scriptstyle\pm375.947}$ & 82.0 & 2.0 & \checkmark & 56.0 & $6.87\!\times\!10^{3}{\scriptstyle\pm 6.00\!\times\!10^{4}}$ & 100.0 & 1.0 & \checkmark & 37.0 & $4.49\!\times\!10^{4}{\scriptstyle\pm 4.39\!\times\!10^{5}}$ & 100.0 \\
   & SafeDiffuser~\cite{xiao2025safediffuser} & 67.0 & \checkmark & 33.0 & $107.749{\scriptstyle\pm190.770}$ & 24.0 & 49.0 & \checkmark & 46.0 & $59.366{\scriptstyle\pm9.082}$ & 0.0 & 13.0 & \checkmark & 71.0 & $70.873{\scriptstyle\pm30.022}$ & 18.0 \\
\midrule
  \multirow{4}{*}{Flow Matching}
   & FM~\cite{yang2026safeflowmatcher} & $97.0^\dagger$ & - & 47.0 & $47.356{\scriptstyle\pm21.188}$ & - & $88.0^\dagger$ & - & 61.0 & $43.229{\scriptstyle\pm22.993}$ & - & $62.0^\dagger$ & - & 61.0 & $36.489{\scriptstyle\pm22.204}$ & - \\
   & SafeFM~\cite{yang2026safeflowmatcher} & 80.0 & \checkmark & 20.0 & $50.885{\scriptstyle\pm24.197}$ & 0.0 & 46.0 & \checkmark & 51.0 & $54.196{\scriptstyle\pm42.148}$ & 1.0 & 23.0 & \checkmark & 57.0 & $148.792{\scriptstyle\pm181.809}$ & 38.0 \\
   & FlowMatcher~\cite{yang2026safeflowmatcher} & $95.0^\dagger$ & - & 40.0 & $84.885{\scriptstyle\pm15.125}$ & - & $90.0^\dagger$ & - & 70.0 & $60.688{\scriptstyle\pm9.721}$ & - & $67.0^\dagger$ & - & 52.0 & $55.324{\scriptstyle\pm8.650}$ & - \\
   & SafeFlowMatcher~\cite{yang2026safeflowmatcher} & 66.0 & \checkmark & 34.0 & $90.772{\scriptstyle\pm20.488}$ & 1.0 & 25.0 & \checkmark & 70.0 & $61.578{\scriptstyle\pm10.628}$ & 0.0 & 24.0 & \checkmark & 56.0 & $54.889{\scriptstyle\pm8.634}$ & 2.0 \\
\midrule
  \multirow{4}{*}{\shortstack[l]{Streaming Flow\\ (Ours)}}
   & StreamingFlow (open-loop) & $94.0^\dagger$ & - & 48.0 & $13.880{\scriptstyle\pm2.623}$ & - & $93.0^\dagger$ & - & 36.0 & $6.426{\scriptstyle\pm1.060}$ & - & $89.0^\dagger$ & - & 46.0 & $3.923{\scriptstyle\pm0.875}$ & - \\
   & StreamingFlow (closed-loop) & $100.0^\dagger$ & - & 45.0 & $12.019{\scriptstyle\pm2.303}$ & - & $94.0^\dagger$ & - & 41.0 & $5.321{\scriptstyle\pm1.215}$ & - & $91.0^\dagger$ & - & 50.0 & $3.119{\scriptstyle\pm0.699}$ & - \\
   & \ssf (open-loop) & \textbf{98.0} & \checkmark & 0.0 & $11.259{\scriptstyle\pm1.744}$ & \textbf{0.0} & 93.0 & \checkmark & 1.0 & $5.049{\scriptstyle\pm0.688}$ & \textbf{0.0} & 76.0 & \checkmark & 18.0 & $3.119{\scriptstyle\pm0.658}$ & \textbf{0.0} \\
   & \ssf (closed-loop) & \textbf{98.0} & \checkmark & 0.0 & $11.504{\scriptstyle\pm2.000}$ & \textbf{0.0} & \textbf{95.0} & \checkmark & \textbf{0.0} & $5.070{\scriptstyle\pm0.967}$ & \textbf{0.0} & \textbf{88.0} & \checkmark & \textbf{4.0} & $3.253{\scriptstyle\pm0.703}$ & \textbf{0.0} \\
\bottomrule
\end{tabular}%
}
\vspace{2pt}\\
\makebox[\textwidth][l]{\footnotesize $^\dagger$ For methods without active safety, the Succ.\ is computed without the safety condition.}
\vspace{-18pt}
\end{table*}

Table~\ref{tab:maze2d_main} reports Maze2D performance across the three maps.
Among methods without a safety mechanism, StreamingFlow (closed-loop) is the strongest on Succ.\ and Sm.\ across all three maps.
Enforcing safety whenever a solution exists leaves \ssf\ at essentially the same Succ.\ and Sm.\ as StreamingFlow, with positive $\bar{h}_{\min}$, zero traps, and zero or near-zero violations in execution, whereas the trajectory-level safe variants drop sharply in Succ., fail to ensure safety in execution, and incur trap discontinuities.
The classical planners, which search the known map, reach a comparable Succ.\ ($85$--$94\%$).
A qualitative visualization of the actual trajectories on the map layout is shown in Appendix~\ref{app:maze2d} (Fig.~\ref{fig:maze_plan_rollout}).

Additionally, by enforcing safety only at the executed step, \ssf reduces the number of optimization solves from $\mathcal{O}(HK)$ to $\mathcal{O}(H)$, cutting optimization-solve time by $\sim\!5.4\times$ and end-to-end latency well below the trajectory-level safe baselines.
A full computational breakdown on Maze2D Large is given in Appendix~\ref{app:computation}.

\subsection{Autonomous Racing: F1TENTH}\label{subsec:f1tenth}
\begin{table*}[h]
\caption{F1TENTH results on the Budapest and Catalunya tracks.}
\label{tab:f1tenth_main}
\centering
\footnotesize
\setlength{\tabcolsep}{3pt}
\resizebox{\textwidth}{!}{%
\begin{tabular}{l l cccccc cccccc}
\toprule
& & \multicolumn{6}{c}{\textbf{Budapest} ($H{=}696;\ K{=}512$)}
  & \multicolumn{6}{c}{\textbf{Catalunya} ($H{=}724;\ K{=}512$)} \\
\cmidrule(lr){3-8} \cmidrule(lr){9-14}
\textbf{Category} & \textbf{Method}
  & Succ. & $\bar h_{\min}{>}0$ & Viol. & Sm. & $t_{\text{Opt}}$\,(s) & $t_{\text{total}}$\,(s)
  & Succ. & $\bar h_{\min}{>}0$ & Viol. & Sm. & $t_{\text{Opt}}$\,(s) & $t_{\text{total}}$\,(s) \\
\midrule
\multirow{3}{*}{Diffusion}
  & Diffuser~\cite{janner2022planning}
    & $79.0^\dagger$ & -- & 80.0 & $19.22{\scriptstyle\pm0.11}$ & -- & 1.84
    & $69.0^\dagger$ & -- & 1.0 & $20.14{\scriptstyle\pm0.11}$ & -- & 1.88 \\
  & Diffuser + CG~\cite{dhariwal2021diffusion}
    & 0.0 & \checkmark & 86.0 & $21.02{\scriptstyle\pm0.12}$ & 0.17 & 2.07
    & 59.0 & \checkmark & 0.0 & $20.15{\scriptstyle\pm0.10}$ & 0.47 & 2.40 \\
  & SafeDiffuser~\cite{xiao2025safediffuser}
    & 0.0 & \checkmark & 80.0 & $19.57{\scriptstyle\pm0.18}$ & 17.88 & 19.97
    & 0.0 & \checkmark & 0.0 & $1.29\!\times\!10^{4}{}^{\ddagger}$ & 19.59 & 21.68 \\
\midrule
\multirow{4}{*}{Flow Matching}
  & FM~\cite{yang2026safeflowmatcher}
    & $47.0^\dagger$ & -- & 30.0 & $5.89{\scriptstyle\pm3.05}$ & -- & 1.76
    & $68.0^\dagger$ & -- & 29.0 & $6.20{\scriptstyle\pm4.15}$ & -- & 1.80 \\
  & SafeFM~\cite{yang2026safeflowmatcher}
    & 0.0 & \checkmark & 0.0 & $1.93\!\times\!10^{3}{}^{\ddagger}$ & 17.72 & 19.72
    & 0.0 & \checkmark & 0.0 & $1.50\!\times\!10^{4}{}^{\ddagger}$ & 19.27 & 21.28 \\
  & FlowMatcher~\cite{yang2026safeflowmatcher}
    & $0.0^\dagger$ & -- & 0.0 & $160.6{\scriptstyle\pm14.1}$ & -- & 1.77
    & $0.0^\dagger$ & -- & 2.0 & $86.5{\scriptstyle\pm3.7}$ & -- & 1.84 \\
  & SafeFlowMatcher~\cite{yang2026safeflowmatcher}
    & 0.0 & \checkmark & 0.0 & $161.1{\scriptstyle\pm13.8}$ & 18.16 & 20.19
    & 0.0 & \checkmark & 1.0 & $5.49\!\times\!10^{3}{}^{\ddagger}$ & 19.73 & 21.78 \\
\midrule
\multirow{2}{*}{\shortstack[l]{Streaming Flow\\(Ours)}}
  & StreamingFlow (open-loop)
    & $77.0^\dagger$ & -- & 97.0 & $0.880{\scriptstyle\pm0.010}$ & -- & 3.23
    & $99.0^\dagger$ & -- & 0.0 & $0.736{\scriptstyle\pm0.045}$ & -- & 3.36 \\
  & \ssf (open-loop)
    & \textbf{73.0} & \checkmark & 20.0 & $0.961{\scriptstyle\pm0.016}$ & 0.71 & 3.96
    & \textbf{99.0} & \checkmark & 0.0 & $0.736{\scriptstyle\pm0.045}$ & 0.49 & 3.99 \\
\bottomrule
\end{tabular}%
}
\vspace{2pt}\\
\makebox[\textwidth][l]{\footnotesize $^\dagger$ For methods without active safety, the Succ.\ is computed without the safety condition.}\\
\makebox[\textwidth][l]{\footnotesize $^\ddagger$ The plans do not follow the track; their waypoints are scattered over the data range.}
\vspace{-18pt}
\end{table*}

To evaluate scalability beyond navigation tasks, we test on the F1TENTH autonomous racing simulator with nonholonomic bicycle dynamics~\cite{kelly2020f1tenth}, using two tracks---Budapest and Catalunya---with substantially longer horizons than Maze2D (Table~\ref{tab:f1tenth_main}).

Table~\ref{tab:f1tenth_main} reports the F1TENTH results. The baselines struggle at these horizon lengths: most flow matching and FlowMatcher variants achieve $0\%$ goal-reaching success on Budapest, and SafeDiffuser attains $0\%$ Succ.\ on both tracks.
Diffuser reaches $88\%$ on Budapest but violates every barrier constraint ($\bar{h}_{\min} \leq 0$).
In contrast, StreamingFlow achieves $77\%$ and $99\%$ success on the two tracks with the smoothest plans under long-horizon streaming generation. \ssf further lowers the violation rate on Budapest from $97\%$ to $20\%$ ($\bar{h}_{\min} > 0$ on every rollout across both tracks) while preserving the goal-reaching of StreamingFlow and comparable smoothness.

\subsection{Locomotion: Hopper}\label{subsec:hopper}

To assess \ssf beyond navigation-style tasks, we evaluate on MuJoCo Hopper~\cite{fu2020d4rl}, a single-leg locomotion system.
Unlike the finite-horizon, goal-reaching tasks of Maze2D and F1TENTH, Hopper is a receding-horizon locomotion task with no terminal goal state, so the height constraint must be enforced continually throughout the rollout.
Here, we encoded it as a velocity-augmented ceiling barrier; because the barrier itself carries velocity, its relative degree collapses to 1, requiring only a single class-$\mathcal{K}$ condition.
The reduction to a single class-$\mathcal{K}$ condition, together with metrics and experimental details, is given in Appendix~\ref{app:hopper}.

\begin{table*}[h]
\centering
\footnotesize
\setlength{\tabcolsep}{4pt}

\begin{minipage}[t]{0.49\textwidth}
\centering
\caption{Hopper with a torso-height constraint.}
\label{tab:hopper}
\resizebox{\textwidth}{!}{%
\begin{tabular}{l ccccc}
\toprule
\multicolumn{6}{c}{\textbf{Hopper} ($H{=}1000$, $K{=}20$)} \\
\midrule
\textbf{Method}
& \textbf{Score}
& $\bar h_{\min}{>}0$
& $t_{\text{Opt}}$ (ms)
& $t_{\text{total}}$ (ms)
& \textbf{Viol.}\,(\%) \\
\midrule
Diffuser~\cite{janner2022planning}              & $71.2^\dagger{\scriptstyle\pm15.2}$ & --  & --  & $393.8$ & -- \\
Diffuser + CG~\cite{dhariwal2021diffusion}      & $17.4{\scriptstyle\pm13.8}$ & $\times$   & $176.9$ & $591.6$  & $80.0$ \\
SafeDiffuser~\cite{xiao2025safediffuser}        & $16.6{\scriptstyle\pm13.7}$ & \checkmark & $597.8$ & $1012.4$ & $85.0$ \\
\midrule
FM~\cite{yang2026safeflowmatcher}               & $56.5^\dagger{\scriptstyle\pm1.9}$  & --  & --  & $244.3$ & -- \\
SafeFM~\cite{yang2026safeflowmatcher}           & $25.6{\scriptstyle\pm18.8}$ & \checkmark & $561.3$ & $817.7$  & $60.0$ \\
FlowMatcher~\cite{yang2026safeflowmatcher}      & $59.4^\dagger{\scriptstyle\pm7.7}$  & --  & --  & $245.2$ & -- \\
SafeFlowMatcher~\cite{yang2026safeflowmatcher}  & $13.5{\scriptstyle\pm5.0}$ & \checkmark & $570.9$ & $829.5$  & $99.0$ \\
\midrule
StreamingFlow                                   & $57.4^\dagger{\scriptstyle\pm5.5}$  & --  & --  & $7.54$  & -- \\
\ssf                                            & $\mathbf{51.4{\scriptstyle\pm3.4}}$  & \checkmark & $\mathbf{0.013}$ & $\mathbf{7.30}$ & $\mathbf{0.0}$ \\
\bottomrule
\end{tabular}%
}
\end{minipage}
\hfill
\begin{minipage}[t]{0.49\textwidth}
\centering
\caption{Warehouse navigation.}
\label{tab:warehouse_main}
\resizebox{\textwidth}{!}{%
\begin{tabular}{l cccccc}
\toprule
\multicolumn{7}{c}{\textbf{Warehouse Navigation} ($H{=}512$, $K{=}256$)} \\
\midrule
\textbf{Method}
  & Succ. & $\bar h_{\min}{>}0$ & Viol. & Sm. & $t_{\text{Opt}}$\,(s) & $t_{\text{total}}$\,(s) \\
\midrule
RRT*~\cite{karaman2011rrtstar}
    & \textbf{56.0} & \checkmark & 0.0 & $0.087{\scriptstyle\pm0.256}$ & -- & $3.35{\scriptstyle\pm2.37}$ \\
A*~\cite{hart1968astar}
    & 41.0 & \checkmark & 0.0 & $0.092{\scriptstyle\pm0.169}$ & -- & $0.36{\scriptstyle\pm0.34}$ \\
\midrule
Diffuser~\cite{janner2022planning}
    & $50.0^\dagger$ & -- & 25.0 & $0.335{\scriptstyle\pm0.129}$ & -- & $0.78{\scriptstyle\pm0.00}$ \\
Diffuser + CG~\cite{dhariwal2021diffusion}
    & 37.0 & $\times$ & 12.0 & $0.395{\scriptstyle\pm0.283}$ & 0.10 & $0.92{\scriptstyle\pm0.00}$ \\
SafeDiffuser~\cite{xiao2025safediffuser}
    & 34.0 & \checkmark & 19.0 & $0.686{\scriptstyle\pm1.037}$ & 6.91 & $7.82{\scriptstyle\pm0.42}$ \\
\midrule
FM~\cite{yang2026safeflowmatcher}
    & $58.0^\dagger$ & -- & 18.0 & $0.222{\scriptstyle\pm0.162}$ & -- & $0.76{\scriptstyle\pm0.00}$ \\
SafeFM~\cite{yang2026safeflowmatcher}
    & 37.0 & \checkmark & 19.0 & $1.160{\scriptstyle\pm1.604}$ & 7.78 & $8.66{\scriptstyle\pm0.67}$ \\
FlowMatcher~\cite{yang2026safeflowmatcher}
    & $45.0^\dagger$ & -- & 13.0 & $0.599{\scriptstyle\pm0.820}$ & -- & $0.76{\scriptstyle\pm0.00}$ \\
SafeFlowMatcher~\cite{yang2026safeflowmatcher}
    & 38.0 & \checkmark & 11.0 & $0.584{\scriptstyle\pm0.750}$ & 3.28 & $4.12{\scriptstyle\pm3.79}$ \\
\midrule
StreamingFlow (open-loop)
    & $57.0^\dagger$ & -- & 15.0 & $0.113{\scriptstyle\pm0.051}$ & -- & $2.40{\scriptstyle\pm0.00}$ \\
StreamingFlow (closed-loop)
    & $54.0^\dagger$ & -- & 17.0 & $0.047{\scriptstyle\pm0.019}$ & -- & $2.51{\scriptstyle\pm0.00}$ \\
\ssf (open-loop)
    & \textbf{51.0} & \checkmark & 7.0 & $0.113{\scriptstyle\pm0.050}$ & $\mathbf{0.09}$ & $2.56{\scriptstyle\pm0.07}$ \\
\ssf (closed-loop)
    & \textbf{51.0} & \checkmark & \textbf{1.0} & $0.045{\scriptstyle\pm0.020}$ & $\mathbf{0.07}$ & $2.58{\scriptstyle\pm0.00}$ \\
\bottomrule
\end{tabular}%
}
\end{minipage}
\vspace{2pt}
\makebox[\textwidth][c]{\footnotesize $^\dagger$ For methods without active safety, the Score and Succ.\  are computed without the safety condition.}
\vspace{-15pt}
\end{table*}

Table~\ref{tab:hopper} reports score, two layers of safety, and per-action latency.
All trajectory-level CBF variants achieve plan-level safety ($\bar h_{\min}{>}0$) yet incur high \emph{execution-level} violation rates ($46$--$85\%$); a certified plan does not survive the closed-loop deviations of underactuated, contact-rich dynamics. \ssf is the only method attaining safety at \emph{both layers simultaneously}---zero execution-time violations at score $51.4$ ($90\%$ of its unsafe counterpart StreamingFlow's $57.4$) and $7.54$\,ms latency, versus SafeDiffuser's $1012.4$\,ms ($134\times$ slower) at a far worse score ($16.6$).
By acting at every emission step, rather than once on the full horizon, the per-step HOCBF-QP filter compensates for closed-loop drift in real time.

\subsection{Warehouse Navigation in Gazebo}\label{subsec:gazebo}

\begin{figure}[h]
\centering
\includegraphics[width=0.55\columnwidth]{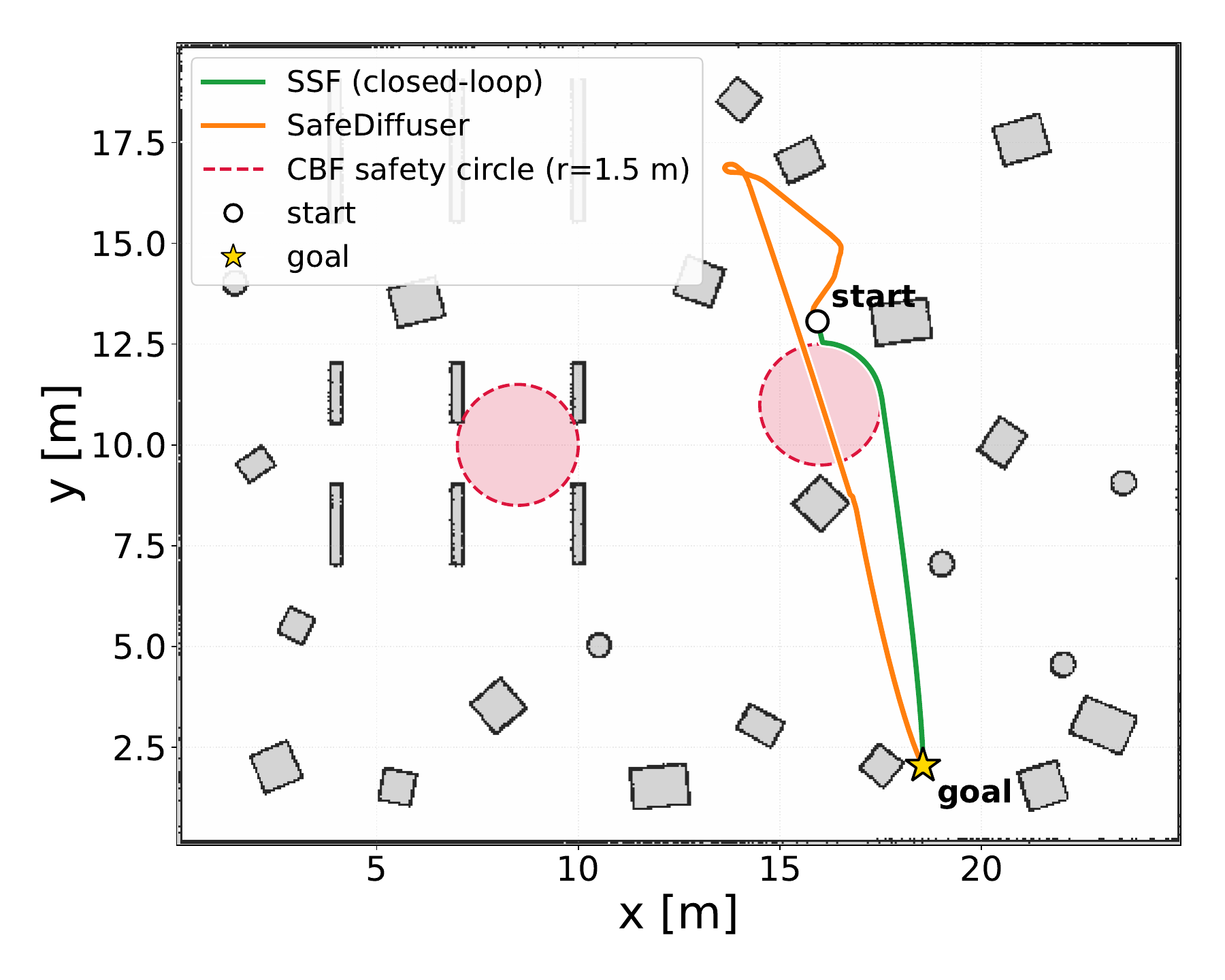}
\vspace{-10px}
\caption{Gazebo warehouse executed trajectories. SafeDiffuser (orange) violates the CBF region (red dashed) during execution; \ssf (green) stays clear.}
\vspace{-10px}
\label{fig:gazebo}
\end{figure}

Table~\ref{tab:warehouse_main} reports the static warehouse benchmark, a ROS2/Gazebo environment that runs the full Nav2 deployment stack and a physics engine; this exercises every method under the closed-loop control loop a physical robot would impose.
All learned methods share a single PD controller and the classical baselines use Nav2 (setup in Appendix~\ref{app:gazebo}).
As in Hopper, a plan that satisfies the constraint need not survive closed-loop execution: SafeDiffuser drifts into the safety region during rollout while \ssf{} stays clear (Figure~\ref{fig:gazebo}).
Among safe generative planners, the streaming family is the strongest on goal-reaching, and in closed-loop execution produces markedly smoother trajectories than the others.
Crucially, \ssf enforces safety whenever a solution exists, almost for free.
In closed loop, it has no violation while executing its plan; the single violation in Table~\ref{tab:warehouse_main} occurs after the planned horizon (Appendix~\ref{app:gazebo}).
Its optimization-solve time is more than an order of magnitude below the trajectory-level safe baselines, giving a lower end-to-end latency than these baselines together with the highest safe success among generative methods.
Against RRT* and A* run within Nav2, a navigation-specialized stack with an explicit cost map, \ssf stays competitive in success with a plain PD controller and no cost map, while running faster than RRT*.

\section{Conclusion and Limitations}\label{sec:conclusion}

We presented \ssf, a start--goal-conditioned planner that streams one-step state updates by integrating a learned vector field in physical time.
Because sampling unfolds in physical time, a constraint on the sampling dynamics is a constraint on the executed state under the system model, making safety enforcement dynamically consistent for a generative planner.
A naive streaming predictor exploits a kinematic shortcut in second-order systems, which we resolve with hierarchical state prediction.
This formulation yields three benefits: discrete-time HOCBFs reduce the number of optimization solves from $\mathcal{O}(HK)$ to $\mathcal{O}(H)$, extend to closed-loop execution for free, and avoid local-trap artifacts.
Across four environments (Maze2D, F1TENTH, Hopper, Warehouse Navigation), \ssf improves execution-time safety, smoothness, and latency over safe baselines while remaining competitive on goal-reaching.

\textbf{Limitations:}
Several limitations remain.
First, because \ssf transports data to data rather than noise to data as in conventional flow matching, it fundamentally requires richer demonstration data; our contracting conditional flows alleviate this requirement but do not remove it.
Second, by enforcing safety through CBFs, \ssf inherits their classical limitations: the quadratic barrier restricts the obstacle geometries handled efficiently, and valid barrier functions and their gains must be specified in advance.
Third, as noted after Theorem~\ref{thm:invariance}, we do not establish recursive feasibility, and feasibility is lost even in a simple setting: on Maze2D Large, the filter occasionally finds no solution, and a few closed-loop rollouts violate the barrier.
Finally, all evaluations are in simulation; real-robot deployment is needed to validate practicality and may expose further sim-to-real gaps.


\clearpage
\acknowledgments{This work was supported by the National Research Foundation of Korea (NRF) grant funded by the Korea government (MSIT) (RS-2026-25504174).}


\bibliography{ref}  

@article{ho2020denoising,
  title={Denoising diffusion probabilistic models},
  author={Ho, Jonathan and Jain, Ajay and Abbeel, Pieter},
  journal={Advances in neural information processing systems},
  volume={33},
  pages={6840--6851},
  year={2020}
}

@inproceedings{lipman2022flow,
title={Flow Matching for Generative Modeling},
author={Yaron Lipman and Ricky T. Q. Chen and Heli Ben-Hamu and Maximilian Nickel and Matthew Le},
booktitle={The Eleventh International Conference on Learning Representations },
year={2023},
}

@inproceedings{fahnestock2024learning,
  title={Learning semantic traversability priors using diffusion models for uncertainty-aware global path planning},
  author={Fahnestock, Ethan and Fuentes, Erick and Osteen, Philip R and Ancha, Siddharth and Roy, Nicholas},
  booktitle={ICRA 2024 Workshop on Resilient Off-road Autonomy},
  year={2024},
  organization={IEEE}
}

@inproceedings{jiang2025anomalies,
  title={Anomalies-by-Synthesis: Anomaly Detection using Generative Diffusion Models for Off-Road Navigation},
  author={Jiang, Sunshine and Ancha, Siddharth and Manderson, Travis and Brandt, Laura and Du, Yilun and Osteen, Philip R and Roy, Nicholas},
  booktitle={2025 IEEE International Conference on Robotics and Automation (ICRA)},
  pages={13320--13329},
  year={2025},
  organization={IEEE}
}

@article{chi2024diffusionpolicy,
	author = {Cheng Chi and Zhenjia Xu and Siyuan Feng and Eric Cousineau and Yilun Du and Benjamin Burchfiel and Russ Tedrake and Shuran Song},
	title ={Diffusion Policy: Visuomotor Policy Learning via Action Diffusion},
	journal = {The International Journal of Robotics Research},
	year = {2024},
}

@inproceedings{prasad2024consistency,
  title={Consistency Policy: Accelerated Visuomotor Policies via Consistency Distillation},
  author={Prasad, Aaditya and Lin, Kevin and Wu, Jimmy and Zhou, Linqi and Bohg, Jeannette},
  booktitle={Robotics: Science and Systems},
  year={2024}
}

@inproceedings{hoeg2025fast,
  author={Høeg, Sigmund H. and Du, Yilun and Egeland, Olav},
  booktitle={2025 IEEE International Conference on Robotics and Automation (ICRA)}, 
  title={Fast Policy Synthesis with Variable Noise Diffusion Models}, 
  year={2025},
  volume={},
  number={},
  pages={4821-4828},
  doi={10.1109/ICRA55743.2025.11127858}}

@inproceedings{jiang2025streaming,
title={Streaming Flow Policy: Simplifying diffusion/flow-matching policies by treating action trajectories as flow trajectories},
author={Sunshine Jiang and Xiaolin Fang and Nicholas Roy and Tom{\'a}s Lozano-P{\'e}rez and Leslie Pack Kaelbling and Siddharth Ancha},
booktitle={9th Annual Conference on Robot Learning},
year={2025},
url={https://openreview.net/forum?id=jnpILGz9gQ}
}

@inproceedings{janner2022planning,
  title={Planning with Diffusion for Flexible Behavior Synthesis},
  author={Janner, Michael and Du, Yilun and Tenenbaum, Joshua and Levine, Sergey},
  booktitle={International Conference on Machine Learning},
  pages={9902--9915},
  year={2022},
  organization={PMLR}
}

@inproceedings{carvalho2023motion,
  title={Motion planning diffusion: Learning and planning of robot motions with diffusion models},
  author={Carvalho, Joao and Le, An T and Baierl, Mark and Koert, Dorothea and Peters, Jan},
  booktitle={2023 IEEE/RSJ International Conference on Intelligent Robots and Systems (IROS)},
  pages={1916--1923},
  year={2023},
  organization={IEEE}
}

@inproceedings{ye2024efficient,
  title={Efficient trajectory forecasting and generation with conditional flow matching},
  author={Ye, Sean and Gombolay, Matthew C},
  booktitle={2024 IEEE/RSJ International Conference on Intelligent Robots and Systems (IROS)},
  pages={2816--2823},
  year={2024},
  organization={IEEE}
}

@inproceedings{
xiao2025safediffuser,
title={SafeDiffuser: Safe Planning with Diffusion Probabilistic Models},
author={Wei Xiao and Tsun-Hsuan Wang and Chuang Gan and Ramin Hasani and Mathias Lechner and Daniela Rus},
booktitle={The Thirteenth International Conference on Learning Representations},
year={2025},
}

@inproceedings{
yang2026safeflowmatcher,
title={{SafeFlowMatcher}: Safe and Fast Planning using Flow Matching with Control Barrier Functions},
author={Jeongyong Yang and Seunghwan Jang and SooJean Han},
booktitle={The Fourteenth International Conference on Learning Representations},
year={2026},
}

@inproceedings{mizuta2024cobl,
  title={Cobl-diffusion: Diffusion-based conditional robot planning in dynamic environments using control barrier and lyapunov functions},
  author={Mizuta, Kazuki and Leung, Karen},
  booktitle={2024 IEEE/RSJ International Conference on Intelligent Robots and Systems (IROS)},
  pages={13801--13808},
  year={2024},
  organization={IEEE}
}

@article{dhariwal2021diffusion,
  title={Diffusion models beat gans on image synthesis},
  author={Dhariwal, Prafulla and Nichol, Alexander},
  journal={Advances in neural information processing systems},
  volume={34},
  pages={8780--8794},
  year={2021}
}

@inproceedings{
ho2021classifierfree,
title={Classifier-Free Diffusion Guidance},
author={Jonathan Ho and Tim Salimans},
booktitle={NeurIPS 2021 Workshop on Deep Generative Models and Downstream Applications},
year={2021},
}

@article{ames2016control,
  title={Control barrier function based quadratic programs for safety critical systems},
  author={Ames, Aaron D and Xu, Xiangru and Grizzle, Jessy W and Tabuada, Paulo},
  journal={IEEE Transactions on Automatic Control},
  volume={62},
  number={8},
  pages={3861--3876},
  year={2016},
  publisher={IEEE}
}

@inproceedings{ames2019control,
  title={Control barrier functions: Theory and applications},
  author={Ames, Aaron D and Coogan, Samuel and Egerstedt, Magnus and Notomista, Gennaro and Sreenath, Koushil and Tabuada, Paulo},
  booktitle={2019 18th European control conference (ECC)},
  pages={3420--3431},
  year={2019},
  organization={IEEE}
}

@inproceedings{Nguyen2016exponential,
  author={Nguyen, Quan and Sreenath, Koushil},
  booktitle={2016 American Control Conference (ACC)}, 
  title={Exponential Control Barrier Functions for enforcing high relative-degree safety-critical constraints}, 
  year={2016},
  volume={},
  number={},
  pages={322-328},
}

@ARTICLE{xiong2023discretehocbf,
  author={Xiong, Yuhan and Zhai, Di-Hua and Tavakoli, Mahdi and Xia, Yuanqing},
  journal={IEEE Transactions on Cybernetics}, 
  title={Discrete-Time Control Barrier Function: High-Order Case and Adaptive Case}, 
  year={2023},
  volume={53},
  number={5},
  pages={3231-3239},
}

@article{fu2020d4rl,
  title={{D4RL}: Datasets for deep data-driven reinforcement learning},
  author={Fu, Justin and Kumar, Aviral and Nachum, Ofir and Tucker, George and Levine, Sergey},
  journal={arXiv preprint arXiv:2004.07219},
  year={2020}
}

@inproceedings{kelly2020f1tenth,
  title     = {{F1TENTH}: An Open-source Evaluation Environment for Continuous Control and Reinforcement Learning},
  author    = {O'Kelly, Matthew and Zheng, Hongrui and Karthik, Dhruv and Mangharam, Rahul},
  booktitle = {NeurIPS 2019 Competition and Demonstration Track},
  year      = {2020}
}

@article{karaman2011rrtstar,
  title   = {Sampling-based Algorithms for Optimal Motion Planning},
  author  = {Karaman, Sertac and Frazzoli, Emilio},
  journal = {The International Journal of Robotics Research},
  volume  = {30},
  number  = {7},
  pages   = {846--894},
  year    = {2011},
  doi     = {10.1177/0278364911406761}
}

@article{hart1968astar,
  title   = {A Formal Basis for the Heuristic Determination of Minimum Cost Paths},
  author  = {Hart, Peter E. and Nilsson, Nils J. and Raphael, Bertram},
  journal = {IEEE Transactions on Systems Science and Cybernetics},
  volume  = {4},
  number  = {2},
  pages   = {100--107},
  year    = {1968},
  doi     = {10.1109/TSSC.1968.300136}
}

@inproceedings{macenski2020marathon2,
  title     = {The Marathon 2: A Navigation System},
  author    = {Macenski, Steve and Mart\'in, Francisco and White, Ruffin and Gin\'es Clavero, Jonatan},
  booktitle = {2020 IEEE/RSJ International Conference on Intelligent Robots and Systems (IROS)},
  year      = {2020},
  pages     = {2718--2725},
  doi       = {10.1109/IROS45743.2020.9341207}
}

@INPROCEEDINGS{gazebo,
  author={Koenig, N. and Howard, A.},
  booktitle={2004 IEEE/RSJ International Conference on Intelligent Robots and Systems (IROS) (IEEE Cat. No.04CH37566)}, 
  title={Design and use paradigms for Gazebo, an open-source multi-robot simulator}, 
  year={2004},
  volume={3},
  number={},
  pages={2149-2154 vol.3},
  doi={10.1109/IROS.2004.1389727}}

@inproceedings{pmlr-v305-jung25d,
  title     = {Joint Model-based Model-free Diffusion for Planning with Constraints},
  author    = {Jung, Wonsuhk and Mishra, Utkarsh Aashu and Arachchige, Nadun Ranawaka and Chen, Yongxin and Xu, Danfei and Kousik, Shreyas},
  booktitle = {9th Annual Conference on Robot Learning},
  year      = {2025}
}

@inproceedings{pmlr-v305-ma25a,
  title     = {Constraint-Aware Diffusion Guidance for Robotics: Real-Time Obstacle Avoidance for Autonomous Racing},
  author    = {Ma, Hao and Bodmer, Sabrina and Carron, Andrea and Zeilinger, Melanie and Muehlebach, Michael},
  booktitle = {9th Annual Conference on Robot Learning},
  year      = {2025}
}

@article{bodmer2026groundinggenerativepoliciesphysics,
  title   = {Grounding Generative Policies in Physics: Optimization-Guided Diffusion for Robot Control},
  author  = {Bodmer, Sabrina and Zurbr{\"u}gg, Ren{\'e} and Portela, Tifanny and Ma, Hao and Didier, Alexandre and Hutter, Marco and Jones, Colin and Zeilinger, Melanie},
  journal = {arXiv preprint arXiv:2606.24208},
  year    = {2026}
}

@inproceedings{bansal2017hamilton,
  title     = {{H}amilton-{J}acobi Reachability: A Brief Overview and Recent Advances},
  author    = {Bansal, Somil and Chen, Mo and Herbert, Sylvia and Tomlin, Claire J.},
  booktitle = {IEEE Conference on Decision and Control},
  year      = {2017}
}

@article{wabersich2021predictive,
  title   = {A Predictive Safety Filter for Learning-Based Control of Constrained Nonlinear Dynamical Systems},
  author  = {Wabersich, Kim P. and Zeilinger, Melanie N.},
  journal = {Automatica},
  volume  = {129},
  pages   = {109597},
  year    = {2021}
}
\newpage
\appendix

\section{Notation}\label{app:notation}
\begin{table}[h]
\caption{Notation.}
\label{tab:notation}
\centering
\footnotesize
\begin{tabular}{ll}
\toprule
\textbf{Symbol} & \textbf{Meaning} \\
\midrule
$\mathbf{x}=(\mathbf{p},\mathbf{v})$ & state: position and velocity \\
$\mathbf{u}$, $\mathcal{U}$ & control input and admissible input set \\
$f, g$ ($\tilde f, \tilde g$) & drift and input terms of the dynamics (of the velocity equation) \\
$T$, $H$, $\Delta t=T/H$ & horizon time, number of steps, sampling time \\
$k$, $\mathcal{T}=\{0,\ldots,H-1\}$ & discrete step and its index set \\
$K$ & sampling steps of a trajectory-level baseline \\
$\mathbf{x}_s$, $\mathbf{x}_g$, $c=(\mathbf{x}_s,\mathbf{x}_g)$ & start, goal, and conditioning \\
$\bm{\tau}$, $\mathcal{D}$ & demonstration and demonstration dataset \\
$F_\theta=(F^p_\theta,F^v_\theta)$ & learned vector field and its position and velocity branches (HSP) \\
$F_{\bm{\tau}}$, $A_{\Delta t}$ & trajectory-conditioned target field and kinematic lift \\
$\sigma_t=\sigma_0e^{-\lambda t}$ & noise scale of the conditional paths, with contracting gain $\lambda$ \\
$\hat{\mathbf{p}}_{k+1\mid k}$, $\hat{\mathbf{v}}_{k+1\mid k}$ & one-step state reference \\
$K_p$, $K_v$; $\mathbf{u}^{\mathrm{PD}}_k$, $\mathbf{u}^{\mathrm{safe}}_k$ & PD gains; nominal and filtered inputs \\
$h_i$, $\mathcal{I}$, $\mathcal{C}$ & barrier functions, their index set, and the safe set \\
$\psi_{i,0},\psi_{i,1},\psi_{i,2}$ & discrete-time HOCBF sequence \\
$\mathcal{C}_{i,0},\mathcal{C}_{i,1},\mathcal{C}_i$ & sets defined by $\psi_{i,0}\ge0$, $\psi_{i,1}\ge0$, and both \\
$\alpha_{i,1},\alpha_{i,2}$ ($k_p,k_v$) & class-$\mathcal{K}$ functions (their linear gains) \\
$\kappa$ & safety margin in the barrier \\
$\Delta\in\{\textsc{open},\textsc{closed}\}$ & planning mode \\
\bottomrule
\end{tabular}
\end{table}

\section{\ssf Planning Pseudocode}\label{appendix:alg}
\begin{algorithm}[h]
\caption{\ssf}
\label{alg:ssf}
\begin{algorithmic}[1]
\Require system dynamics~\eqref{eq:dynamics}; learned fields $F^p_\theta, F^v_\theta$; start and goal $\mathbf{x}_s, \mathbf{x}_g$; step size $\Delta t$; planning horizon $H$; PD gains $K_p, K_v$; barrier functions $\{h_i\}_{i\in\mathcal{I}}$ with $\{(\alpha_{i,1},\alpha_{i,2})\}_{i\in\mathcal{I}}$; planning mode $\Delta\in\{\textsc{open},\textsc{closed}\}$.
\State Initialize $\mathbf{x}_0$ as $\mathbf{x}_s$.
\For{$k \in \mathcal{T}$}
    \If{$\Delta$ == \textsc{closed}}
        \State Observe $\mathbf{x}_k = (\mathbf{p}_k,\mathbf{v}_k)$.
    \EndIf
    \State Compute $(\hat{\mathbf{p}}_{k+1 \mid k},\hat{\mathbf{v}}_{k+1 \mid k})$ via \eqref{eq:HSP_pos}--\eqref{eq:HSP_vel}.
    \State Compute $\mathbf{u}^{\mathrm{PD}}_k$ via \eqref{eq:pd_control}.
    \State Solve optimization problem in \eqref{eq:hocbf_qcqp} to obtain $\mathbf{u}^{\mathrm{safe}}_k$.
    \State Propagate the state via the system dynamics~\eqref{eq:dynamics}:
    \State \hspace{\algorithmicindent}
    $\mathbf{x}_{k+1} \gets \mathbf{x}_k + \Delta t\big(f(\mathbf{x}_k)+g(\mathbf{x}_k)\mathbf{u}^{\mathrm{safe}}_k\big)$
    \If{$\Delta$ == \textsc{closed}}
        \State Robot agent tracks $\mathbf{x}_{k+1}$ for any controller.
    \EndIf
\EndFor
\State \Return $\{\mathbf{x}_{k}\}_{k\in\mathcal{T}\cup\{H\}}$
\end{algorithmic}
\end{algorithm}

\section{Proof of Theorem~\ref{thm:invariance} (Forward Invariance)}\label{app:proof_thm1}
The argument follows~\cite{xiong2023discretehocbf}, specialized to relative degree two; we state the result for several constraints satisfied by a single applied input.
Fix $i\in\mathcal{I}$. We show $\mathbf{x}_k\in\mathcal{C}_i$ for all $k\ge 0$ by induction on $k$; the base case holds because $\mathbf{x}_0\in\bigcap_{j\in\mathcal{I}}\mathcal{C}_j\subseteq\mathcal{C}_i$.
Suppose $\mathbf{x}_k\in\mathcal{C}_i$, i.e., $\psi_{i,0}(\mathbf{x}_k)\ge 0$ and $\psi_{i,1}(\mathbf{x}_k)\ge 0$, and let $\mathbf{x}_{k+1}$ be the successor of $\mathbf{x}_k$ under~\eqref{eq:dt_dynamics} with the applied input $\mathbf{u}_k$.
Since $\alpha_{i,1}$ and $\alpha_{i,2}$ are class-$\mathcal{K}$ functions with $\alpha(s)<s$ for $s>0$, we have $s-\alpha(s)\ge 0$ for all $s\ge 0$.
By~\eqref{eq:dt_psi2}, condition~\eqref{eq:dt_hocbf_condition} at the applied input gives
\begin{equation*}
    \psi_{i,1}(\mathbf{x}_{k+1}) \ge \psi_{i,1}(\mathbf{x}_k)-\alpha_{i,2}\big(\psi_{i,1}(\mathbf{x}_k)\big) \ge 0,
\end{equation*}
so $\mathbf{x}_{k+1}\in\mathcal{C}_{i,1}$.
By~\eqref{eq:dt_psi1}, $\psi_{i,1}(\mathbf{x}_k)\ge 0$ gives
\begin{equation*}
    \psi_{i,0}(\mathbf{x}_{k+1}) \ge \psi_{i,0}(\mathbf{x}_k)-\alpha_{i,1}\big(\psi_{i,0}(\mathbf{x}_k)\big) \ge 0,
\end{equation*}
so $\mathbf{x}_{k+1}\in\mathcal{C}_{i,0}$; here $\psi_{i,0}(\mathbf{x}_{k+1})$ is the value at the actual successor, since $h_i$ depends only on position and $\mathbf{p}_{k+1}=\mathbf{p}_k+\Delta t\,\mathbf{v}_k$ does not depend on $\mathbf{u}_k$.
Hence $\mathbf{x}_{k+1}\in\mathcal{C}_i$, and by induction $\mathbf{x}_k\in\mathcal{C}_i$ for all $k\ge 0$.
Since $i$ was arbitrary, $\mathbf{x}_k\in\bigcap_{i\in\mathcal{I}}\mathcal{C}_i\subseteq\bigcap_{i\in\mathcal{I}}\mathcal{C}_{i,0}=\mathcal{C}$ for all $k\ge 0$, where the last equality is~\eqref{eq:safe_set}.
\hfill\qed

\section{Proof of Theorem~\ref{thm:marginal_matching} (Marginal Matching)}\label{appendix:proof_marginal}
Fix condition $c=(\mathbf{x}_s,\mathbf{x}_g)$, let $\Phi_t^{\bm{\tau}}(\boldsymbol{\xi}) \triangleq \mathbf{x}_t^{\bm{\tau}} = \bm{\tau}(t)+\sigma_t A_{\Delta t}\boldsymbol{\xi}$, and define the trajectory-conditioned distribution $\mu_t^{\bm{\tau}}=(\Phi_t^{\bm{\tau}})_\# \mathcal{N}(0,I_n)$. The start--goal-conditioned target marginal is $\mu_t^c \triangleq \int \mu_t^{\bm{\tau}} \, p_{\mathcal D}(d\bm{\tau}\mid c) = \mathbb{E}_{\bm{\tau} \sim p_{\mathcal D}(\cdot \mid c)}[\mu_t^{\bm{\tau}}]$.

We show that the target marginal curve $\{\mu_t^c\}_{t\in[0,T]}$ satisfies the continuity equation induced by the optimal vector field $F^*$ in the weak sense:
\begin{equation}\label{eq:marginal_continuity}
    \partial_t \mu_t^c + \nabla\cdot\left(\mu_t^c F^*(\cdot, t \mid c)\right) = 0.
\end{equation}
Equivalently, for every $\varphi\in C^1(\mathbb{R}^{2n})$, it suffices to prove
\begin{equation}\label{eq:weak_goal}
    \frac{d}{dt}\int \varphi(\mathbf{x})\,d\mu_t^c(\mathbf{x}) = \int \nabla\varphi(\mathbf{x})^\top F^*(\mathbf{x},t\mid c)\,d\mu_t^c(\mathbf{x}).
\end{equation}
Consider a fixed demonstration $\bm{\tau}$.
Differentiating the noise schedule $\sigma_t=\sigma_0 e^{-\lambda t}$ in~\eqref{eq:traj_cond_state_sample} gives $\dot{\sigma}_t=-\lambda\sigma_t$, so the time derivative of~\eqref{eq:traj_cond_state_sample} is
\begin{equation}
    \dot{\mathbf{x}}_t^{\bm{\tau}}=\dot{\bm{\tau}}(t)-\lambda\sigma_tA_{\Delta t}\boldsymbol{\xi}=\dot{\bm{\tau}}(t)-\lambda A_{\Delta t}\bigl(\mathbf{p}_{\mathbf{x}_t^{\bm{\tau}}}-\mathbf{p}_{\bm{\tau}}(t)\bigr)=F_{\bm{\tau}}(\mathbf{x}_t^{\bm{\tau}},t).
\end{equation}
For any test function $\varphi\in C^1(\mathcal{X})$, the pushforward definition $\mu_t^{\bm{\tau}}=(\Phi_t^{\bm{\tau}})_\# \mathcal{N}(0,I_n)$ implies
\begin{equation}
	\int \varphi(\mathbf{x})\,d\mu_t^{\bm{\tau}}(\mathbf{x}) = \mathbb{E}_{\boldsymbol{\xi}\sim\mathcal{N}(0,I_n)}\left[\varphi(\Phi_t^{\bm{\tau}}(\boldsymbol{\xi}))\right].
\end{equation}
Differentiating this with respect to $t$ and applying the chain rule yields
\begin{equation}
    \frac{d}{dt} \int \varphi(\mathbf{x})\,d\mu_t^{\bm{\tau}}(\mathbf{x})
    \int \nabla\varphi(\mathbf{x})^\top F_{\bm{\tau}}(\mathbf{x},t) \,d\mu_t^{\bm{\tau}}(\mathbf{x}).
\end{equation}
Averaging over $\bm{\tau}\sim p_{\mathcal D}(\cdot\mid c)$ yields
\begin{equation}\label{eq:avg_conditional}
    \frac{d}{dt}\int \varphi(\mathbf{x})\,d\mu_t^c(\mathbf{x})=\int \nabla\varphi(\mathbf{x})^\top \mathbb{E}\!\left[F_{\bm{\tau}}(\mathbf{x},t)\mid \mathbf{x},t,c\right]\,d\mu_t^c(\mathbf{x}).
\end{equation}
The right-hand side of~\eqref{eq:avg_conditional} involves the conditional expectation $\mathbb{E}[F_{\bm{\tau}}(\mathbf{x},t)\mid \mathbf{x},t,c]$.
Since the conditional flow matching loss~\eqref{eq:cfm_loss} is a pointwise least-squares regression problem, its minimizer satisfies
    $F^*(\mathbf{x},t\mid c)=\mathbb{E}\!\left[F_{\bm{\tau}}(\mathbf{x},t)\mid \mathbf{x},t,c\right]$,$\mu_t^c$-a.e.
Substituting this into~\eqref{eq:avg_conditional} proves \eqref{eq:weak_goal}, and $\mu_t^c$ satisfies \eqref{eq:marginal_continuity}, concluding the proof.
\hfill\qed

Note that this proof is different from \cite{jiang2025streaming} because the contraction term is different from that of \cite{jiang2025streaming}, which has singular covariance, i.e., $\mathrm{rank}(A_{\Delta t} A_{\Delta t}^\top)=n$.

\section{QCQP Structure of the Discrete-Time HOCBF Constraint}\label{appendix:qcqp}

We show that the discrete-time HOCBF safety filter used in this work becomes a QCQP when the barrier functions are quadratic and the class-$\mathcal{K}$ functions are linear as described in Appendix~\ref{appendix:experiment}. We focus on the position-based quadratic barrier functions in~\eqref{eq:unified_barrier} used in our experiments. This choice is motivated by the fact that such barriers are a common form for collision avoidance and have relative degree two. If velocity is included in the barrier function, the relative degree becomes one rather than two; in this case, a standard CBF is sufficient and an HOCBF is not required.

At time step $k$, the state $\mathbf{x}_k=(\mathbf{p}_k,\mathbf{v}_k)$ is fixed and the optimization variable is the current control input $\mathbf{u}_k$. Let $\tilde f_k \triangleq \tilde f(\mathbf{p}_k,\mathbf{v}_k)$ and  $\tilde g_k \triangleq \tilde g(\mathbf{p}_k,\mathbf{v}_k)$. Under the Euler discretization of the second-order dynamics,
\begin{equation}
\begin{gathered}
    \mathbf{p}_{k+1|k} = \mathbf{p}_k+\Delta t\,\mathbf{v}_k, \\
    \mathbf{v}_{k+1|k} = \mathbf{v}_k + \Delta t \big( \tilde f_k+\tilde g_k\mathbf{u}_k \big).
\end{gathered}
\end{equation}
Since the barrier depends only on position, the input first appears in the two-step-ahead position prediction:
\begin{align}
    \mathbf{p}_{k+2|k}(\mathbf{u}_k)
    &= \mathbf{p}_{k+1|k} + \Delta t\,\mathbf{v}_{k+1|k} = \mathbf{p}_k + 2\Delta t\,\mathbf{v}_k + \Delta t^2\tilde f_k + \Delta t^2\tilde g_k\mathbf{u}_k \triangleq \mathbf{a}_k+B_k\mathbf{u}_k,
\end{align}
where $\mathbf{a}_k \triangleq \mathbf{p}_k + 2\Delta t\,\mathbf{v}_k + \Delta t^2\tilde f_k$ and $B_k \triangleq \Delta t^2\tilde g_k$. Thus, the two-step-ahead position is affine in $\mathbf{u}_k$.
Now define $h_{i,k} \triangleq h_i(\mathbf{p}_k)$, $h_{i,k+1} \triangleq h_i(\mathbf{p}_{k+1|k})$, and $h_{i,k+2}(\mathbf{u}_k) \triangleq h_i(\mathbf{p}_{k+2|k}(\mathbf{u}_k))$. Because $\mathbf{p}_{k+2|k}(\mathbf{u}_k)=\mathbf{a}_k+B_k\mathbf{u}_k$ and $h_i$ is quadratic, we have
\begin{align}
    h_{i,k+2}(\mathbf{u}_k)
    &= (\mathbf{a}_k+B_k\mathbf{u}_k-\mathbf{c}_i)^\top Q_i (\mathbf{a}_k+B_k\mathbf{u}_k-\mathbf{c}_i) -1-\kappa \notag\\
    &= \mathbf{u}_k^\top M_{i,k}\mathbf{u}_k + \ell_{i,k}^\top \mathbf{u}_k + r_{i,k},
\end{align}
where $M_{i,k} \triangleq B_k^\top Q_i B_k$, $\ell_{i,k} \triangleq 2B_k^\top Q_i(\mathbf{a}_k-\mathbf{c}_i)$, and $r_{i,k} \triangleq (\mathbf{a}_k-\mathbf{c}_i)^\top Q_i(\mathbf{a}_k-\mathbf{c}_i) -1-\kappa$.

For linear class-$\mathcal{K}$ functions, the relative degree two discrete-time HOCBF condition \eqref{eq:dt_hocbf_condition} can be written as
\begin{align}
    \psi_{i,2}(\mathbf{x}_k,\mathbf{u}_k)
    &= h_{i,k+2}(\mathbf{u}_k) -2h_{i,k+1} +h_{i,k} \notag\\
    &\quad +\gamma_{i,1} \big( h_{i,k+1}-h_{i,k} \big) + \gamma_{i,2} \big( h_{i,k+1}-h_{i,k} + \gamma_{i,1}h_{i,k} \big) \\
    &= h_{i,k+2}(\mathbf{u}_k) + (\gamma_{i,1}+\gamma_{i,2}-2)h_{i,k+1} + (1-\gamma_{i,1})(1-\gamma_{i,2})h_{i,k}.\notag
\end{align}
Substituting the quadratic expression for $h_{i,k+2}(\mathbf{u}_k)$ gives
\begin{equation}
    \psi_{i,2}(\mathbf{x}_k,\mathbf{u}_k) = \mathbf{u}_k^\top M_{i,k}\mathbf{u}_k + \ell_{i,k}^\top \mathbf{u}_k + \beta_{i,k},
\end{equation}
where $\beta_{i,k} \triangleq r_{i,k} + (\gamma_{i,1}+\gamma_{i,2}-2)h_{i,k+1} + (1-\gamma_{i,1})(1-\gamma_{i,2})h_{i,k}$.
Therefore, each HOCBF constraint has the quadratic form, and we can write the safety filter as follows:
\begin{equation}\label{eq:app_hocbf_qcqp}
\begin{aligned}
    \mathbf{u}_k^{\mathrm{safe}} = \argmin_{\mathbf{u}_k} \quad& \|\mathbf{u}_k-\mathbf{u}_k^{\mathrm{nom}}\|_2^2 \\
    \mathrm{s.t.} \quad& \mathbf{u}_k^\top M_{i,k}\mathbf{u}_k + \ell_{i,k}^\top \mathbf{u}_k + \beta_{i,k} \ge 0, \qquad \forall i\in\mathcal I, \\
    & \mathbf{u}_k\in\mathcal U.
\end{aligned}
\end{equation}
If the admissible input set is given by box constraints, $\mathcal U=\{\mathbf{u}: \mathbf{u}_{\min}\le \mathbf{u}\le \mathbf{u}_{\max}\}$, then \eqref{eq:app_hocbf_qcqp} is a QCQP with a quadratic objective, quadratic HOCBF constraints, and linear input constraints.

\section{Hierarchical State Prediction Ablation: Avoiding the Kinematic Shortcut}
\label{app:hsp_ablation}
\begin{figure}[h]
    \centering
    \includegraphics[width=0.8\linewidth]{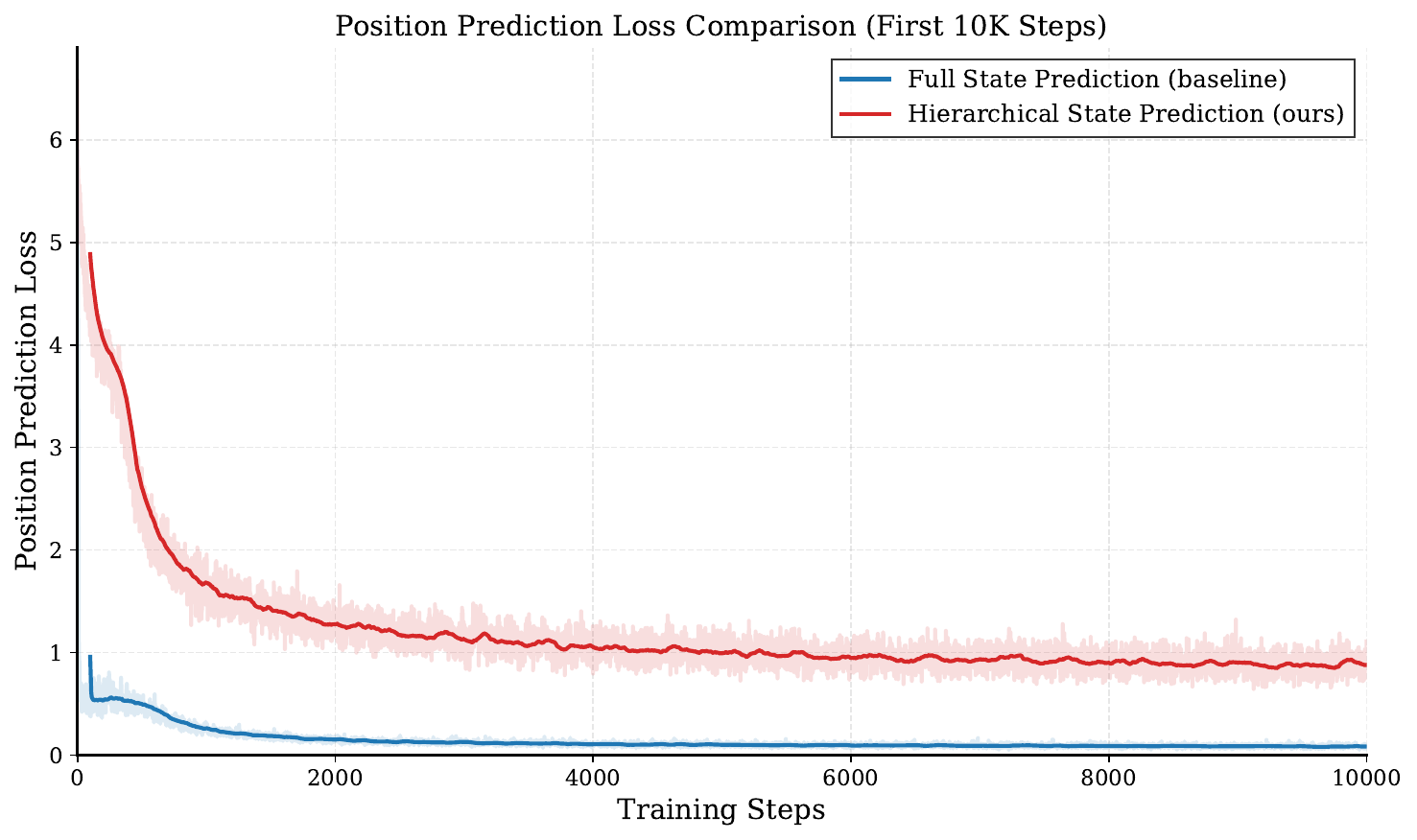}
    \caption{Position prediction loss during training. The FSP quickly reaches near-zero loss by exploiting the kinematic shortcut, whereas HSP converges more slowly to a higher plateau.}
    \label{fig:oaf_pos_loss}
\end{figure}
We ablate the proposed HSP, described in Section~\ref{subsec:hsp}, against a full state prediction (FSP) that directly predicts $(\hat{\mathbf{p}}_{t+\Delta t}, \hat{\mathbf{v}}_{t+\Delta t})$ from the full current state $(\mathbf{p}_t,\mathbf{v}_t)$.

Figure~\ref{fig:oaf_pos_loss} plots the position prediction loss $\|\hat{\mathbf{p}}_{t+\Delta t}-\mathbf{p}^{\star}_{t+\Delta t}\|_2^2$ during training, where $\hat{\mathbf{p}}_{t+\Delta t}$ denotes the predicted next position produced by the learned vector field $F_\theta$ and $\mathbf{p}^{\star}_{t+\Delta t}$ denotes the ground-truth position from the demonstration dataset $\mathcal{D}$.
The HSP decreases slowly and saturates around $0.8$, while the FSP quickly converges to near-zero loss, approximately $0.06$.

This gap does \emph{not} indicate better planning. The FSP can minimize the loss by exploiting what we refer to as a kinematic shortcut. As a result, the next position is predicted almost deterministically from the current velocity, with little reliance on start--goal conditioning.

Table~\ref{tab:hsp_ablation} reports goal-reaching success across all three maps under open- and closed-loop execution. Because success requires reaching within a tight tolerance of the goal, small accumulated errors lead to failure.

\begin{table}[h]
\caption{Ablation on HSP and FSP for the kinematic shortcut: goal-reaching success (\%) across Maze2D maps, open- and closed-loop, without the safety filter. Evaluated over $1{,}000$ rollouts.}
\label{tab:hsp_ablation}
\centering
\footnotesize
\setlength{\tabcolsep}{4pt}
\begin{tabular}{l cc cc cc}
\toprule
& \multicolumn{2}{c}{\textbf{Umaze ($H{=}128$)}} & \multicolumn{2}{c}{\textbf{Medium ($H{=}256$)}} & \multicolumn{2}{c}{\textbf{Large ($H{=}384$)}} \\
\cmidrule(lr){2-3} \cmidrule(lr){4-5} \cmidrule(lr){6-7}
\textbf{Method} & Open & Closed & Open & Closed & Open & Closed \\
\midrule
FSP & 80.2 & 35.0 & 51.1 & 10.8 & 39.6 & 5.8 \\
HSP (ours) & \textbf{94.8} & \textbf{99.7} & \textbf{92.4} & \textbf{91.9} & \textbf{88.6} & \textbf{89.7} \\
\bottomrule
\end{tabular}
\end{table}

Despite its lower training loss, FSP degrades at execution time, particularly as the horizon $H$ increases: shortcut-dependent position predictions accumulate errors over successive steps, causing rapid drift from the goal (e.g., $5.8\%$ success on Large, closed-loop). HSP, by predicting position from $(\mathbf{p}_t,t,\mathbf{x}_s,\mathbf{x}_g)$ alone---without access to $\mathbf{v}_t$---remains robust and achieves substantially higher success across all maps and execution modes.

\paragraph{Training without noise.}
Setting $\sigma_0=0$ places every training state on a demonstration, so the target of~\eqref{eq:cfm_loss} becomes the demonstration velocity and training reduces to next-state regression with the same loss.
Table~\ref{tab:sigma0} compares this variant with $\sigma_0=0.005$ on Maze2D Large; the two perform comparably.

\begin{table}[h]
\caption{Training noise $\sigma_0$ on Maze2D Large (closed loop, safety filter on; mean $\pm$ std over three training seeds, $100$ rollouts each; violations counted at $\kappa=0$).}
\label{tab:sigma0}
\centering
\footnotesize
\begin{tabular}{l ccc}
\toprule
$\sigma_0$ & Goal & Viol. & Succ. \\
\midrule
$0$     & $81.3{\scriptstyle\pm4.2}$ & $2.0{\scriptstyle\pm1.7}$ & $81.3{\scriptstyle\pm4.2}$ \\
$0.005$ & $85.0{\scriptstyle\pm4.0}$ & $3.0{\scriptstyle\pm0.0}$ & $85.0{\scriptstyle\pm4.0}$ \\
\bottomrule
\end{tabular}
\end{table}

\section{Justifying Hierarchical State Prediction}\label{app:hsp_justification}

The HSP ablation (Appendix~\ref{app:hsp_ablation}, Table~\ref{tab:hsp_ablation}) shows that the FSP attains lower training loss yet far lower closed-loop success than HSP. This appendix explains \emph{why}, as a four-link causal chain.

\begin{enumerate}\itemsep1pt
\item[\textbf{L1}] The kinematic shortcut $\mathbf{p}_{t+\Delta t}\!\approx\!\mathbf{p}_t+\Delta t\,\mathbf{v}_t$ exists in the demonstration data.
\item[\textbf{L2}] FSP \emph{actually learned} that shortcut, not merely that it could.
\item[\textbf{L3}] FSP \emph{ignores} the goal as a consequence.
\item[\textbf{L4}] The shortcut \emph{causally} drives closed-loop failure, which HSP avoids by construction.
\end{enumerate}

All experiments use D4RL Maze2D (Umaze/Medium/Large), seed $42$. FSP is the monolithic predictor of Appendix~\ref{app:hsp_ablation}; HSP is the proposed factorized predictor whose position branch has no $\mathbf{v}$ input.

\subsection{The shortcut exists, and FSP learned it (links L1, L2)}

For every demonstration transition we compute the forward-Euler residual $\mathbf{r}=\mathbf{p}_{t+\Delta t}-(\mathbf{p}_t+\Delta t\,\mathbf{v}_t)$ (physical maze units), and evaluate the trained predictors' next-position MSE in the same units. We make two predictions:
\begin{itemize}[leftmargin=*,topsep=2pt,itemsep=2pt]
    \item \textbf{Data side.} Mean $\|\mathbf{r}\|^2$ and its fraction of $\mathrm{Var}(\mathbf{p}_{t+\Delta t})$ are $\approx 0$ (else there is no shortcut to exploit).
    \item \textbf{Model side.} FSP MSE $\approx$ floor (it fits the shortcut as tightly as the data allows); HSP MSE $\gg$ floor (it discarded the shortcut).
\end{itemize}

\begin{table}[h]
\caption{E1/E1b. Data residual floor and per-model next-position MSE (physical units). The shortcut is $5$--$6$ orders below $\mathrm{Var}(\mathbf{p}_{t+\Delta t})$; FSP sits at the floor, HSP $28$--$71\times$ above it.}
\label{tab:e1}
\centering\footnotesize
\begin{tabular}{l ccc cc}
\toprule
Map & floor $\|\mathbf{r}\|^2$ & frac & FSP MSE & HSP MSE & FSP\,/\,HSP\,/\,floor \\
\midrule
Umaze  & $2.69\mathrm{e}{-5}$ & $2.1\mathrm{e}{-5}$ & $1.31\mathrm{e}{-5}$ & $7.49\mathrm{e}{-4}$ & $0.49$ / $27.8$ \\
Medium & $1.50\mathrm{e}{-5}$ & $3.7\mathrm{e}{-6}$ & $5.81\mathrm{e}{-6}$ & $8.11\mathrm{e}{-4}$ & $0.39$ / $53.9$ \\
Large  & $1.27\mathrm{e}{-5}$ & $1.3\mathrm{e}{-6}$ & $4.76\mathrm{e}{-6}$ & $8.99\mathrm{e}{-4}$ & $0.37$ / $70.6$ \\
\bottomrule
\end{tabular}
\end{table}

Both predictions hold on every map ($n\!\approx\!1$--$4{\times}10^{6}$ transitions). FSP's MSE is at, and slightly below, the forward-Euler floor: with $\mathbf{v}$ as input it can fit the data's semi-implicit-Euler kinematics, which is $O(\Delta t^2)$ tighter than forward Euler; HSP, lacking $\mathbf{v}$, cannot. The HSP/floor ratio grows with horizon ($28\!\to\!54\!\to\!71$), as expected when longer rollouts amplify the discarded residual.

\subsection{Functional form of the shortcut (link L2)}

If FSP implements the shortcut, its position-output Jacobian w.r.t.\ velocity must match the closed-form slope $1/(\Delta t_{\mathrm{flow}}\,\mathrm{vel\_weight})$ analytically implied by forward Euler plus the input normalization. We finite-difference $\mathbf{J}=\partial\mathbf{v}_{\mathrm{pos}}/\partial\mathbf{v}$ over $2000$ probes. \textbf{Prediction:} FSP $\mathbf{J}$ diagonal, slope $=$ analytical value, $R^2(\mathbf{v}_{\mathrm{pos}}\!\leftrightarrow\!\mathbf{v})\!\approx\!1$; HSP $|\mathbf{J}|_{\max}\!=\!0$ (architectural).

\begin{table}[h]
\caption{FSP matches the analytical kinematic-shortcut slope to $<1\%$ on every map; the map is essentially linear ($R^2\!\ge\!0.996$, diagonal fraction $>0.994$). HSP $|\mathbf{J}|_{\max}\!=\!0$ exactly, since its position branch has no $\mathbf{v}$ input.}
\label{tab:e3}
\centering\footnotesize
\begin{tabular}{l cc cc c}
\toprule
Map & analytic slope & measured slope & rel.\ err & $R^2_{\min}$ & diag.\ frac \\
\midrule
Umaze  & $(4.696,5.121)$ & $(4.666,5.088)$ & ${<}0.7\%$ & $0.9959$ & $0.9972$ \\
Medium & $(4.805,4.643)$ & $(4.785,4.643)$ & ${<}0.5\%$ & $0.9959$ & $0.9983$ \\
Large  & $(5.870,4.094)$ & $(5.856,4.080)$ & ${<}0.4\%$ & $0.9967$ & $0.9948$ \\
\bottomrule
\end{tabular}
\end{table}

To three significant figures, FSP is the closed-form linear map from $\mathbf{v}$ to $\hat{\mathbf{p}}$.

\subsection{FSP ignores the goal (link L3)}
If FSP plans via the velocity shortcut, its prediction should be (near-)invariant to the commanded goal, whereas HSP's must vary with it. For probe states held fixed we sweep the goal over a bank of $64$ and measure the prediction variance $\bar S$, normalized by the data-side goal-discriminative variance $D_{\mathrm{var}}$ (kNN estimate over the probes with upper-median $D_{\mathrm{var}}$). \textbf{Prediction:} $\bar S_{\mathrm{FSP}}\!\ll\!\bar S_{\mathrm{HSP}}$, and replacing the goal perturbs HSP far more than FSP.

\begin{table}[h]
\caption{E2. Goal sensitivity, normalized by data-side discriminative variance, and goal-permutation change. FSP is two-to-three orders of magnitude less goal-sensitive than HSP on every map.}
\label{tab:e2}
\centering\footnotesize
\setlength{\tabcolsep}{5pt}
\begin{tabular}{l ccc c}
\toprule
Map & $\bar S_{\mathrm{HSP}}$ & $\bar S_{\mathrm{FSP}}$ & $\bar S_{\mathrm{FSP}}/\bar S_{\mathrm{HSP}}$ & perm.\ ratio \\
\midrule
Umaze  & $0.695$ & $1.54\mathrm{e}{-3}$ & $2.2\mathrm{e}{-3}$ & $0.037$ \\
Medium & $0.173$ & $2.13\mathrm{e}{-4}$ & $1.2\mathrm{e}{-3}$ & $0.026$ \\
Large  & $0.144$ & $1.68\mathrm{e}{-4}$ & $1.2\mathrm{e}{-3}$ & $0.021$ \\
\bottomrule
\end{tabular}
\end{table}

\begin{figure}[H]
\centering
\includegraphics[width=\columnwidth]{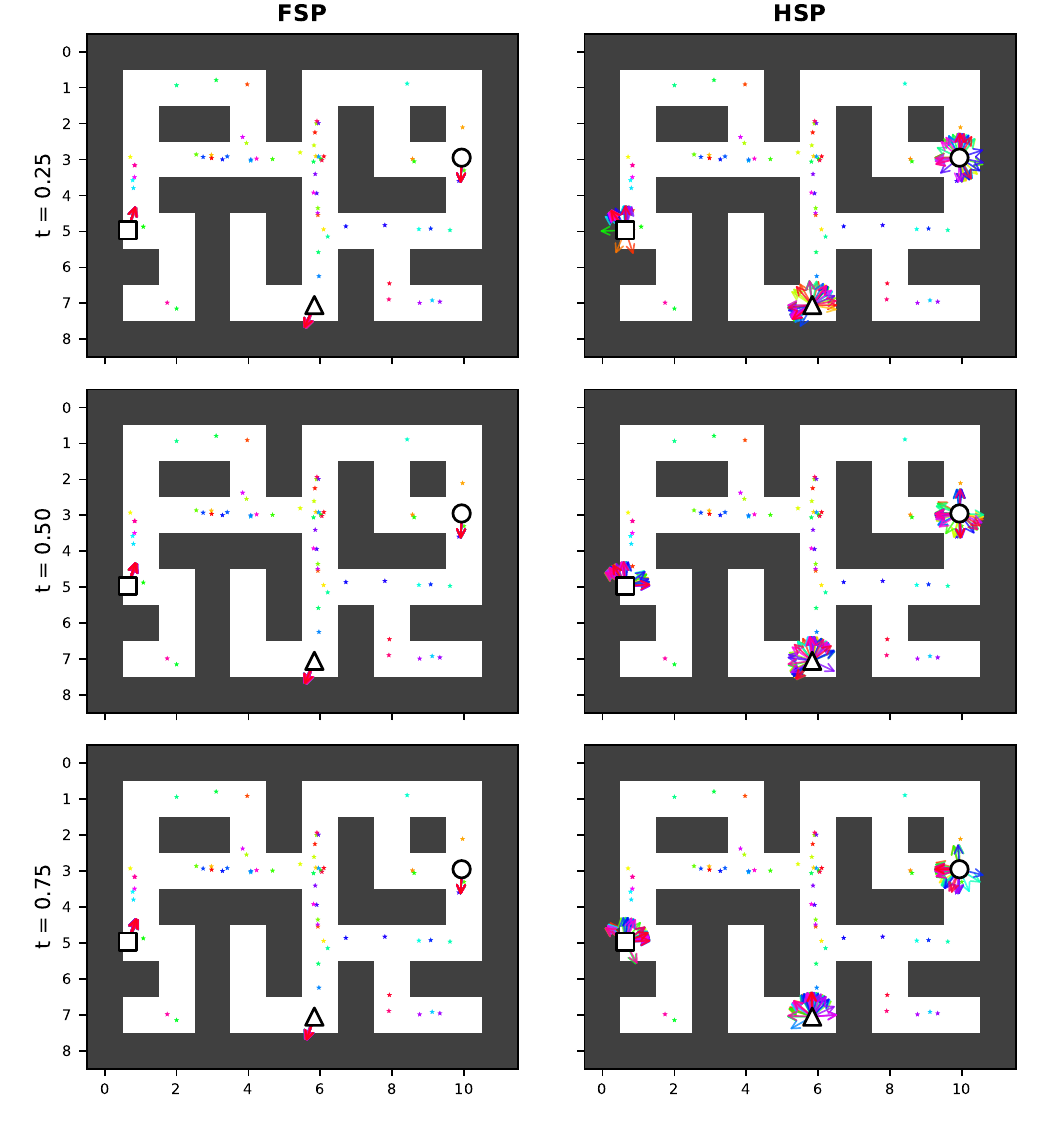}
\vspace{-20pt}
\caption{Visual instantiation of E2 on Maze2D Large. For three goal-discriminative probes ($\bullet,\blacksquare,\blacktriangle$; deterministic selection from the upper-median-$D_{\mathrm{var}}$ pool), we sweep the same $64$-goal bank used in Table~\ref{tab:e2} and plot the unit-direction one-step displacement $(\hat{\mathbf{p}}-\mathbf{p})/\|\hat{\mathbf{p}}-\mathbf{p}\|$, color-matched to goal index. Columns: FSP vs HSP; rows: flow time $t\in\{0.25,0.5,0.75\}$ (both networks consume $t$ via a sinusoidal step embedding). A tight clump of arrows per probe means the model ignores the goal channel; a multi-color fan means it attends to it. FSP arrows clump regardless of the goal; HSP arrows fan toward their respective goals, with the fan most pronounced as $t\!\to\!1$.}
\label{fig:e2_goalsweep}
\end{figure}

Figure~\ref{fig:e2_goalsweep} makes this concrete on Maze2D Large: sweeping the goal leaves FSP's one-step direction in a single goal-invariant bundle, while HSP's fans toward each respective goal.

\subsection{Causal closed-loop sweep (link L4)}

We inject $\boldsymbol{\varepsilon}\!\sim\!\mathcal{N}(0,\sigma^2 I)$ into the \emph{predictor's} velocity input each step ($\sigma\!\in\![0,0.10]$), leaving the true dynamics propagation unchanged, across $\{$open,closed$\}\times\{$filter on,off$\}$. \textbf{Prediction:} FSP success collapses with $\sigma$ while HSP stays flat, the effect being velocity-channel-specific and amplified in closed loop.

\begin{table}[h]
\caption{E4. Success@$\varepsilon{=}0.5$ at $\sigma{=}0\!\to\!0.10$, over $15$/$15$/$10$ rollouts per cell (Umaze/Medium/Large). HSP is flat everywhere. FSP collapses in \emph{closed loop already at $\sigma{=}0$}, worsening with horizon ($0.33\!\to\!0.07\!\to\!0.00$); the gap is far smaller open-loop, isolating the failure to closed-loop velocity feedback.}
\label{tab:e4}
\centering\footnotesize
\begin{tabular}{l cc cc}
\toprule
 & \multicolumn{2}{c}{closed, filter off} & \multicolumn{2}{c}{open, filter off} \\
\cmidrule(lr){2-3}\cmidrule(lr){4-5}
Map & FSP $0{\to}.1$ & HSP $0{\to}.1$ & FSP $0{\to}.1$ & HSP $0{\to}.1$ \\
\midrule
Umaze  & $.33{\to}.20$ & $1.0{\to}1.0$ & $.87{\to}.93$ & $.93{\to}.93$ \\
Medium & $.07{\to}.07$ & $1.0{\to}1.0$ & $.47{\to}.53$ & $1.0{\to}1.0$ \\
Large  & $.00{\to}.10$ & $1.0{\to}1.0$ & $.60{\to}.60$ & $.90{\to}.90$ \\
\bottomrule
\end{tabular}
\end{table}

In closed loop the realized state drifts from the nominal rollout; for FSP this drift, entering through the velocity channel, corrupts the position prediction via the shortcut even with \emph{no} injected noise, so explicit $\sigma$ adds little further degradation (FSP is already at its closed-loop floor). HSP's position branch is invariant to that channel and stays flat. This explains the closed-loop collapse in Table~\ref{tab:hsp_ablation} (Appendix~\ref{app:hsp_ablation}): the position branch of HSP does not see $\mathbf{v}$, including the drift that feedback control introduces.

The shortcut captures only the first-order term $\Delta t\,\mathbf{v}_t$; the residual $\mathbf{r}=\mathbf{p}_{t+\Delta t}-(\mathbf{p}_t+\Delta t\,\mathbf{v}_t)$ is the acceleration content $\tfrac12 \Delta t^2 \mathbf{a}_t+\mathcal{O}(\Delta t^3)$. Two consecutive positions therefore identify $\mathbf{v}_t$ (the average velocity), not $\mathbf{v}_{t+\Delta t}$: the latter requires learning the increment $\mathbf{v}_{t+\Delta t}-\mathbf{v}_t=\Delta t\,\mathbf{a}_t$, which is exactly what $F^v_\theta$ provides.

\vspace{0pt}

\section{Multimodality}\label{app:multimodality}
\begin{figure}[h]
\centering
\includegraphics[width=0.6\linewidth]{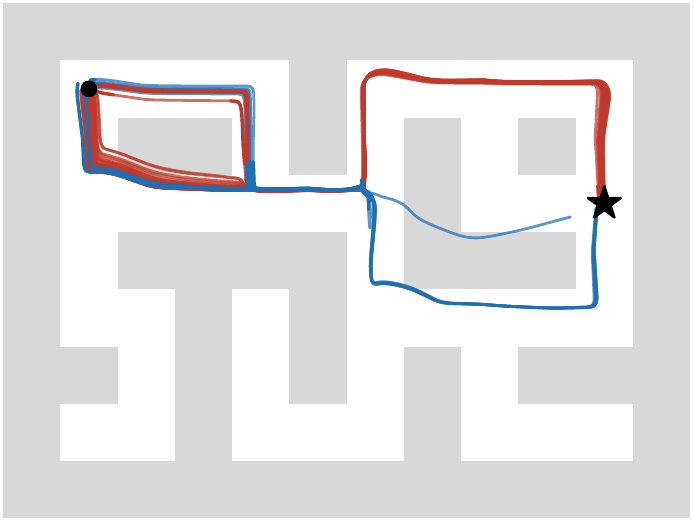}
\caption{Plans from one start ($\bullet$) to one goal ($\star$) on Maze2D Large; small samples of $\mathbf{p}_0$ near $\mathbf{p}_s$ yield two routes, shown in two colors.}
\label{fig:multimodal}
\end{figure}

This section shows experimentally that \ssf produces multimodal plans by adding small noise to the initial position, as in StreamingFlow\cite{jiang2025streaming}.
At test time, we sample $\mathbf{p}_0\sim\mathcal{N}(\mathbf{p}_s,\sigma_{\text{test}}^2 I)$ and keep the conditioning $(\mathbf{x}_s,\mathbf{x}_g)$ unchanged; with $\sigma_{\text{test}}=0$, the plan is unique.
Figure~\ref{fig:multimodal} shows $50$ open-loop plans on Maze2D Large for one start--goal pair with $\sigma_{\text{test}}=0.02$, without the safety filter: the plans split into two routes that pass a wall block on opposite sides ($43$ and $7$ plans).

Noise at the start yields several plans, but this alone does not make the planner multimodal: if the direction of the perturbation decided the route, the plans would only carry the input noise.
We therefore test whether the route can be predicted from the perturbation, on the $19$ groups of $50$ plans that produce more than one route (two start--goal pairs, three training seeds, and $\sigma_{\text{test}}\in\{0.005,0.02,0.05,0.1\}$).

In $15$ of these groups, the route cannot be predicted from the direction of the perturbation.
Plans that take different routes start together and split partway through the plan: the distance between them grows to $9.3$ times its initial value (median), against $2.9$ times for plans on the same route, and reaches half of its maximum at $36\%$ of the horizon.
The perturbation may bias the route, but the route stays unpredictable from its direction: the flow turns small differences in the start into a choice between routes, so the noise yields distinct routes rather than shifted copies of one plan.

\section{Experiment Details}\label{appendix:experiment}

We document the simulation environment, planning hyperparameters, and barrier function configuration for each experimental setting.
Source code and configurations are included in the supplementary material.

\subsection{Common Setup}\label{app:common}

\paragraph{Hardware.}
All experiments run on a single workstation with an AMD EPYC 9354 CPU and
an NVIDIA RTX 4090 (24\,GB) GPU.

\paragraph{Unified barrier form.}
Maze2D, F1TENTH, and Warehouse navigation all use the same super-ellipse barrier
\begin{equation}\label{eq:unified_barrier}
    h_i(\mathbf{p}) \;=\;
    \Bigl(\tfrac{p_x - c_{x,i}}{r_{x,i}}\Bigr)^{n}
    \;+\; \Bigl(\tfrac{p_y - c_{y,i}}{r_{y,i}}\Bigr)^{n}
    \;-\;(1+\kappa),
\end{equation}
where $(c_{x,i},c_{y,i})$ is the obstacle center, $(r_{x,i},r_{y,i})$ its semi-axes, $n\!\ge\!2$ the order, and $\kappa\!>\!0$ a robust safety margin.
With $r_{x,i}\!=\!r_{y,i}\!=\!r_i$ and $n\!=\!2$ this reduces to the quadratic form $h_i = (\mathbf{p}-\mathbf{c}_i)^\top Q_i (\mathbf{p}-\mathbf{c}_i) - 1 - \kappa$ with $Q_i = r_i^{-2}\,I_2$. Per-task obstacle geometries are visualized in the corresponding subsections below; Hopper uses a different barrier form documented in Section~\ref{app:hopper}.

\paragraph{HOCBF parameterization.}
The discrete-time HOCBF (Section~\ref{subsec:cbf}) is parameterized by class-$\mathcal{K}$ gains $(k_p, k_v)$ in
\begin{align*}\label{eq:appendix_hocbf}
  \psi_0 &= h(\mathbf{p}_k),\\
  \psi_1 &= h(\mathbf{p}_{k+1}) - (1-k_p)\,h(\mathbf{p}_k)\;\ge\; 0,\\
  \psi_2 &= h(\mathbf{p}_{k+2}) - (2-k_p-k_v)\,h(\mathbf{p}_{k+1}) + (1-k_p)(1-k_v)\,h(\mathbf{p}_k)\;\ge\; 0.
\end{align*}
The gains $(k_p, k_v)$ are the discrete-time counterparts of linear class-$\mathcal{K}$ functions $\alpha_1(s)=k_p s$ and $\alpha_2(s)=k_v s$ with $k_p, k_v > 0$; this linear choice is the special case of the HOCBF also known as the exponential CBF~\cite{Nguyen2016exponential}.
Per-environment gains and $\kappa$ are given below.

\paragraph{Metrics.}
Across all tasks we report:
\begin{itemize}[itemsep=0pt,topsep=0pt]
  \item \emph{Succ.}(\%): fraction of rollouts that reach the goal,
        $\|\mathbf{p}_T-\mathbf{p}_g\|_2 < \epsilon$, without violation.
        For methods without active safety, Succ.\ counts goal-reaching only, since no barrier constraint is enforced.
  \item \emph{Viol.}(\%): fraction of rollouts whose executed trajectory enters a barrier region ($h_i<0$) at any step;
        in Hopper, the torso rises above $1.5$\,m.
  \item $\bar h_{\min}{>}0$: \checkmark when, in every rollout, the plan satisfies $h_i\!\ge\!0$ at every waypoint
        (plan-level indicator); not reported for methods without active safety.
  \item \emph{Sm.}: acceleration smoothness of the plan,
        $\tfrac{1}{H}\sum_t\|\dot{\mathbf{v}}_{t+\Delta t}-\dot{\mathbf{v}}_t\|$ (Warehouse: Appendix~\ref{app:gazebo}).
  \item \emph{Trap}(\%): fraction of rollouts with discontinuous safety
        corrections.
  \item $t_{\text{Opt}}$, $t_{\text{total}}$: optimization time and end-to-end latency per generated trajectory (per action in Hopper).
\end{itemize}

\paragraph{Goal and safe success of every method.}
The main tables report Succ.\ as safe success for methods with active safety and as goal-reaching for the others; Table~\ref{tab:full_results} reports both for every method.

\begin{table}[h]
\caption{Goal-reaching rate (G) and safe success (S, goal reached without violation) of every method, in \%. Violations use $\kappa=0.01$ in Maze2D and F1TENTH and $\kappa=0$ in Warehouse. --: not evaluated.}
\label{tab:full_results}
\centering
\footnotesize
\setlength{\tabcolsep}{3pt}
\resizebox{\textwidth}{!}{%
\begin{tabular}{l cc cc cc cc cc cc}
\toprule
& \multicolumn{6}{c}{\textbf{Maze2D}} & \multicolumn{4}{c}{\textbf{F1TENTH}} & \multicolumn{2}{c}{\textbf{Warehouse}} \\
\cmidrule(lr){2-7} \cmidrule(lr){8-11} \cmidrule(lr){12-13}
& \multicolumn{2}{c}{Umaze} & \multicolumn{2}{c}{Medium} & \multicolumn{2}{c}{Large} & \multicolumn{2}{c}{Budapest} & \multicolumn{2}{c}{Catalunya} & & \\
\textbf{Method} & G & S & G & S & G & S & G & S & G & S & G & S \\
\midrule
RRT*~\cite{karaman2011rrtstar}             & 96 & 91 & 95 & 94 & 92 & 86 & -- & -- & -- & -- & 56 & 56 \\
A*~\cite{hart1968astar}                    & 100 & 93 & 96 & 85 & 89 & 85 & -- & -- & -- & -- & 41 & 41 \\
\midrule
Diffuser~\cite{janner2022planning}         & 94 & 62 & 94 & 46 & 74 & 25 & 79 & 0 & 69 & 69 & 50 & 35 \\
Diffuser + CG~\cite{dhariwal2021diffusion} & 54 & 36 & 2 & 2 & 4 & 1 & 86 & 0 & 59 & 59 & 44 & 37 \\
SafeDiffuser~\cite{xiao2025safediffuser}   & 94 & 67 & 92 & 49 & 69 & 13 & 79 & 0 & 0 & 0 & 47 & 34 \\
\midrule
FM~\cite{yang2026safeflowmatcher}              & 97 & 53 & 88 & 33 & 62 & 25 & 47 & 20 & 68 & 47 & 58 & 45 \\
SafeFM~\cite{yang2026safeflowmatcher}          & 99 & 80 & 93 & 46 & 65 & 23 & 0 & 0 & 0 & 0 & 51 & 37 \\
FlowMatcher~\cite{yang2026safeflowmatcher}     & 95 & 60 & 90 & 22 & 67 & 28 & 0 & 0 & 0 & 0 & 45 & 37 \\
SafeFlowMatcher~\cite{yang2026safeflowmatcher} & 97 & 66 & 88 & 25 & 68 & 24 & 0 & 0 & 0 & 0 & 45 & 38 \\
\midrule
StreamingFlow (open-loop)   & 94 & 51 & 93 & 58 & 89 & 46 & 77 & 0 & 99 & 99 & 57 & 45 \\
StreamingFlow (closed-loop) & 100 & 55 & 94 & 53 & 91 & 43 & -- & -- & -- & -- & 54 & 41 \\
\ssf (open-loop)            & 98 & 98 & 94 & 93 & 84 & 76 & 77 & 73 & 99 & 99 & 57 & 51 \\
\ssf (closed-loop)          & 98 & 98 & 95 & 95 & 88 & 88 & -- & -- & -- & -- & 51 & 51 \\
\bottomrule
\end{tabular}%
}
\end{table}

\paragraph{Filter settings and solver statistics.}
Table~\ref{tab:solver} lists the settings of \ssf\ and its closed-loop solver statistics for each benchmark.
Each per-step problem is solved by SLSQP, which fails only on Maze2D Large.

\begin{table}[h]
\caption{Filter settings and closed-loop solver statistics of \ssf. Solved: steps at which the filter solves its optimization problem; Failed: steps at which SLSQP reports failure; At bound: steps at which the applied input reaches the input bound (\% of all steps).}
\label{tab:solver}
\centering
\footnotesize
\setlength{\tabcolsep}{4pt}
\resizebox{\textwidth}{!}{%
\begin{tabular}{l cccc c c ccc}
\toprule
\textbf{Benchmark} & $\sigma_0$ & $\lambda$ & $H$ & $\Delta t$ (s) & PD $(K_p,K_v)$ & Input bound & Solved (\%) & Failed (\%) & At bound (\%) \\
\midrule
Maze2D Umaze  & $0.005$ & $0.1$ & $128$ & $0.01$ & $(1,1)$ & $[-1,1]^2$ & $12.92$ & $0$ & $19.95$ \\
Maze2D Medium & $0.005$ & $0.1$ & $256$ & $0.01$ & $(1,1)$ & $[-1,1]^2$ & $11.95$ & $0$ & $17.20$ \\
Maze2D Large  & $0.005$ & $0.1$ & $384$ & $0.01$ & $(1,1)$ & $[-1,1]^2$ & $10.72$ & $0.125$ & $15.64$ \\
Hopper        & $10^{-4}$ & $2.0$ & $1000$ & $0.008$ & -- & $[-1,1]^3$ & $0.42$ & $0$ & $3.84$ \\
Warehouse     & $0.005$ & $0.1$ & $512$ & $0.05$ & $(2.0,0.2)$ & $\|\mathbf{v}\|\le 0.5$\,m/s & $10.31$ & $0$ & $46.06$ \\
\bottomrule
\end{tabular}%
}
\end{table}

\paragraph{Open-loop versus closed-loop evaluation.}
The goal-reaching tasks (Maze2D, F1TENTH, Warehouse navigation) are
\emph{start--goal-conditioned}: a plan fixes both endpoints,
$\mathbf{x}_0 = \mathbf{x}_s$ and $\mathbf{x}_H = \mathbf{x}_g$ (via inpainting
for diffusion baselines). Replanning at an intermediate step $k$ would require
re-conditioning on $\mathbf{x}_k$---neither endpoint---over a remaining horizon
$H-k \neq H$, outside the fixed-horizon formulation these baselines are trained
for; we therefore evaluate them open-loop. Hopper, a receding-horizon task with
no terminal goal, conditions on the current state alone, so replanning is native
and all methods run closed-loop. \ssf streams one executed state per step and
supports both settings.

\subsection{Maze2D}\label{app:maze2d}

\begin{figure}[h]
\centering
\begin{subfigure}{0.18\textwidth}
  \centering
  \includegraphics[width=\linewidth]{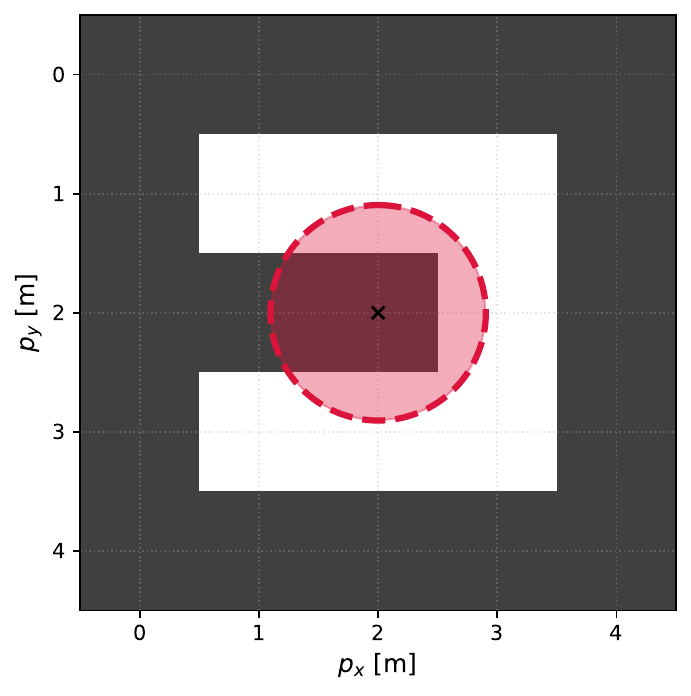}
  \vspace{-8px}
  \caption{Umaze ($5\!\times\!5$)}
  \label{fig:maze_umaze}
\end{subfigure}
\hfill
\begin{subfigure}{0.288\textwidth}
  \centering
  \includegraphics[width=\linewidth]{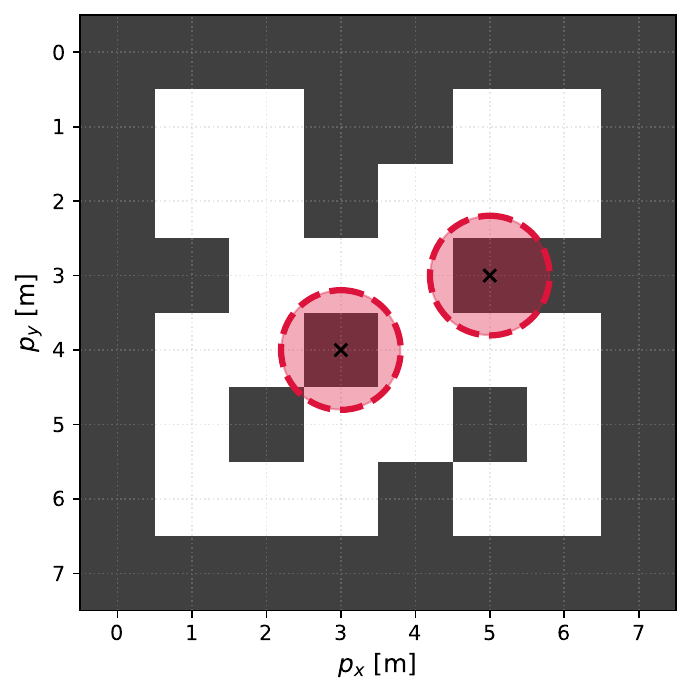}
  \vspace{-13px}
  \caption{Medium ($8\!\times\!8$)}
  \label{fig:maze_medium}
\end{subfigure}
\hfill
\begin{subfigure}{0.432\textwidth}
  \centering
  \includegraphics[width=\linewidth]{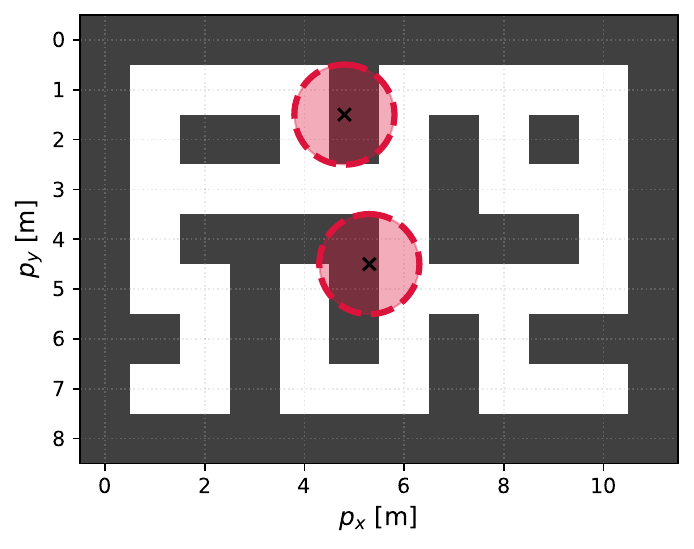}
  \vspace{-20px}
  \caption{Large ($9\!\times\!12$)}
  \label{fig:maze_large}
\end{subfigure}
\caption{Maze2D obstacle layouts for the three D4RL maps. Red regions
indicate the safety-barrier obstacles enforced via
Eq.~\eqref{eq:unified_barrier}; remaining cells form the free space from
which start--goal pairs are uniformly sampled.}
\label{fig:maze_obstacles}
\end{figure}

\paragraph{Environment.}
D4RL \texttt{maze2d-\{umaze,medium,large\}-v1}~\cite{fu2020d4rl}.
Evaluation is conducted over 100 rollouts per map with uniform-random
start--goal sampling from empty cells. Obstacle layouts for the three
maps are shown in Figure~\ref{fig:maze_obstacles}.

\paragraph{Planning horizon and sampling steps.}
\begin{center}\footnotesize
\setlength{\tabcolsep}{4pt}
\begin{tabular}{lccc}
\toprule
& Umaze & Medium & Large \\
\midrule
Map grid        & $5\!\times\!5$ & $8\!\times\!8$ & $9\!\times\!12$ \\
$H$ (horizon)   & $128$ & $256$ & $384$ \\
$K$ (sampling)  & $64$  & $256$ & $256$ \\
\bottomrule
\end{tabular}
\end{center}
\vspace{-2px}
SafeFlowMatcher requires $K+1$ steps due to its two-phase planning structure~\cite{yang2026safeflowmatcher}.

\paragraph{Barrier and HOCBF parameters.}
Eq.~\eqref{eq:unified_barrier} with $n\!=\!2$ and $\kappa\!=\!0.01$ for \ssf\ and for evaluation; the trajectory-level safe baselines use $\kappa\!=\!0.1$ in their filters.
For \ssf, HOCBF gains are $(k_p, k_v) = (0.15, 0.08)$.
The filter's model integrates velocity before position (semi-implicit Euler), so the input already enters $\mathbf{p}_{k+1|k}$; for each obstacle, the QCQP enforces $\psi_{i,2}\ge 0$ together with $h_i(\mathbf{p}_{k+1|k})\ge 0.015$, under the input bound $[-1,1]^2$.
Goal threshold $\epsilon = 0.5$ (D4RL sparse-reward radius).

\paragraph{Classical planners.}
A* and RRT* plan on the known map with the barrier obstacles inflated by $0.05$, and their paths are tracked by the same PD controller.

\begin{figure}[h]
    \centering
    \includegraphics[width=0.6\columnwidth]{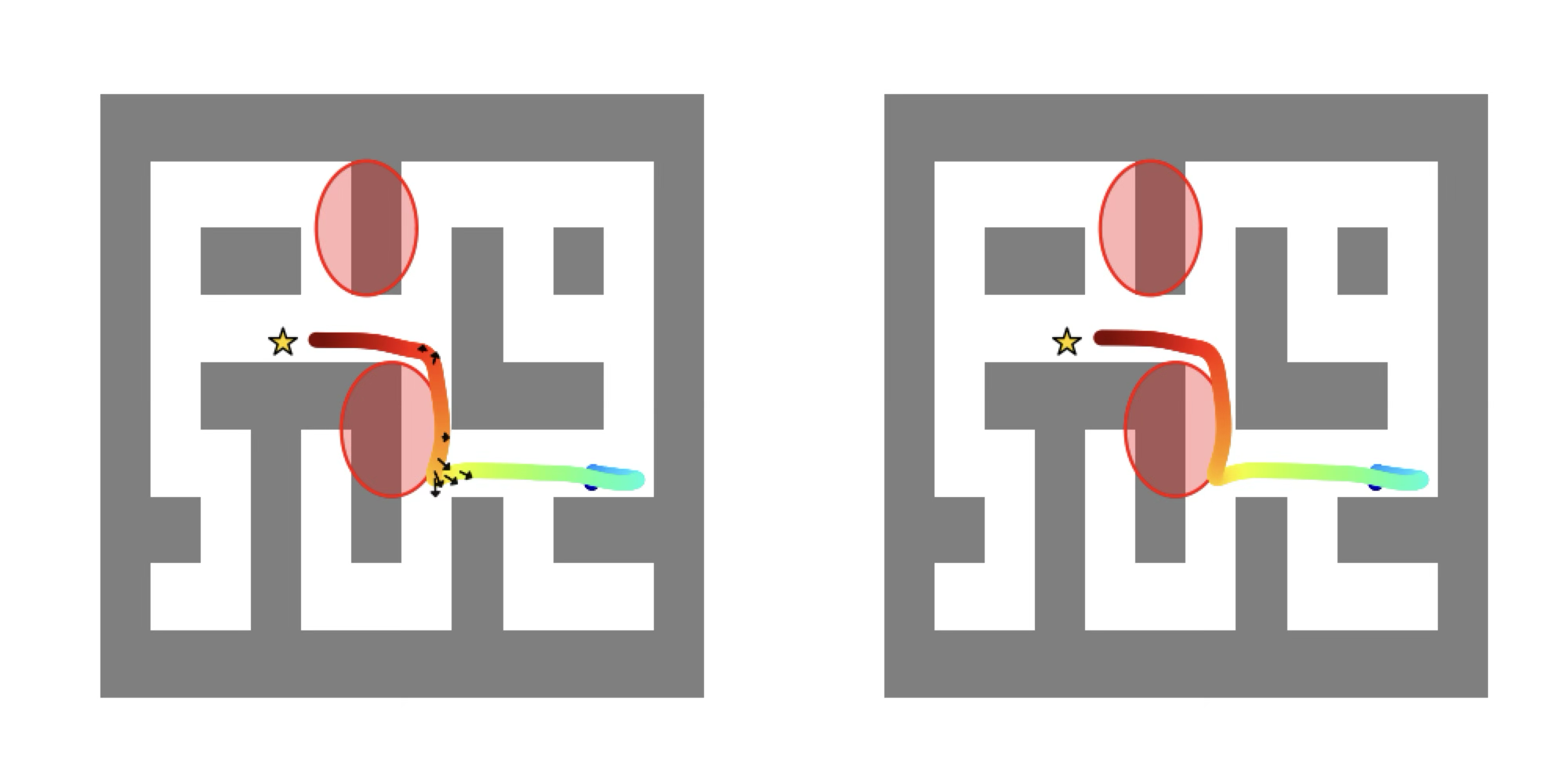}
    \caption{Plan (left) and rollout (right) on Maze2D Large. $\star$ denotes the
    goal $\mathbf{x}_g$; the colored line is the trajectory from start (blue) to
    end (red). Black arrows visualize step-wise HOCBF safety corrections.}
    \label{fig:maze_plan_rollout}
\end{figure}

\paragraph{Qualitative example.}
Figure~\ref{fig:maze_plan_rollout} shows an example plan and the resulting
closed-loop rollout on Maze2D Large.

\paragraph{Computational Efficiency}\label{app:computation}

We measure the cost of generating one trajectory on Maze2D Large
(AMD EPYC 9354 + RTX 4090). Table~\ref{tab:computation} reports total
neural-network forward time (NN), optimization solve time ($t_{\text{Opt}}$),
end-to-end latency ($t_{\text{total}}$), per-step FLOPs, and the number of
optimization solves (\#Opt).

\begin{table}[h]
\caption{Computational efficiency on Maze2D Large. NN, $t_{\text{Opt}}$, and
$t_{\text{total}}$ are totals for generating one full trajectory; FLOPs are
reported per emitted step.}
\label{tab:computation}
\centering
\footnotesize
\resizebox{0.75\columnwidth}{!}{%
\begin{tabular}{l cccc c}
\toprule
\textbf{Method} & NN (ms) & $t_{\text{Opt}}$ (ms) & $t_{\text{total}}$ (ms) & MFLOPs/step & \#Opt \\
\midrule
Diffuser & 921.6 & -- & 921.6 & 771.1 & - \\
Diffuser + CG & 972.8 & 3865.6 & 4838.4 & 771.1 & - \\
SafeDiffuser & 972.8 & 9113.6 & 10086.4 & 771.1 & $H \times K$ \\
\midrule
FM & 896.0 & -- & 896.0 & 771.1 & - \\
SafeFM & 998.4 & 9139.2 & 10137.6 & 771.1 & $H \times K$ \\
FlowMatcher & 896.0 & -- & 896.0 & 771.1 & - \\
SafeFlowMatcher & 972.8 & 9164.8 & 10137.6 & 771.1 & $H \times K$ \\
\midrule
StreamingFlow & 1843.2 & -- & 1843.2 & 120.9 & - \\
\ssf & 1843.2 & 1689.6 & 3532.8 & 120.9 & $H$ \\
\bottomrule
\end{tabular}
}
\end{table}

Trajectory-level safe baselines apply CBF-QP corrections to all $H$ states at
every sampling step, requiring $\mathcal{O}(HK)$ solves for a single trajectory;
\ssf solves a single HOCBF-QCQP per emitted step, totaling $\mathcal{O}(H)$.
Consequently, \ssf reduces the optimization solve time $t_{\text{Opt}}$ by
roughly $5.4\times$ relative to the trajectory-level safe baselines---the
empirical ratio falls short of the theoretical factor of $K$ because the
baselines benefit from batched QP solving---and this is the primary source of
its end-to-end latency advantage. \ssf also incurs the lowest per-step FLOPs
($6.4\times$ below the baselines); its higher NN wall-clock time comes from
unbatched, sequential streaming forward passes.

\subsection{F1TENTH}\label{app:f1tenth}

\begin{figure}[h]
\centering
\begin{subfigure}{0.48\textwidth}
  \centering
  \includegraphics[width=\linewidth]{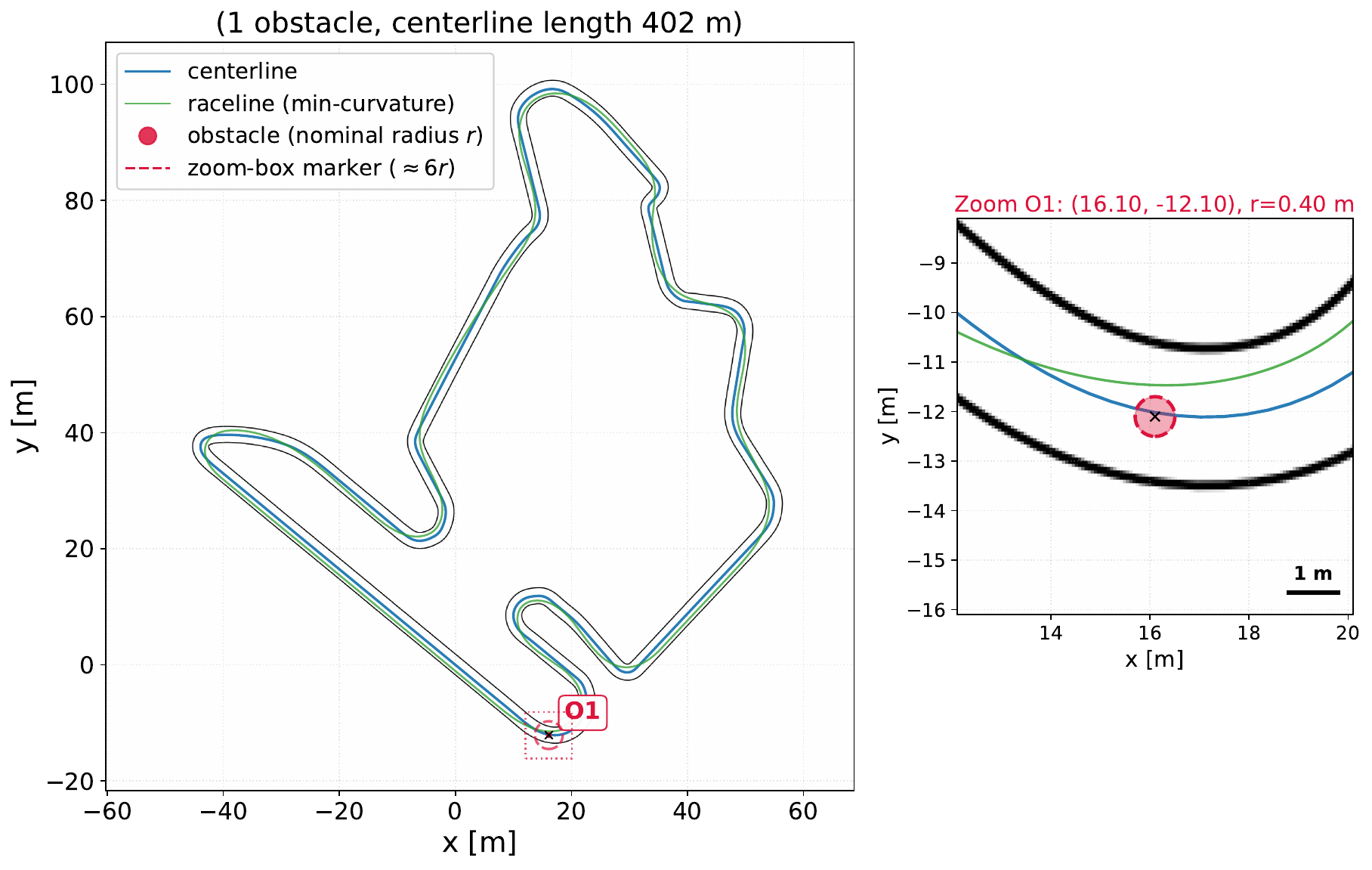}
  \caption{Budapest (1 obstacle).}
  \label{fig:f1tenth_budapest}
\end{subfigure}
\hfill
\begin{subfigure}{0.48\textwidth}
  \centering
  \includegraphics[width=\linewidth]{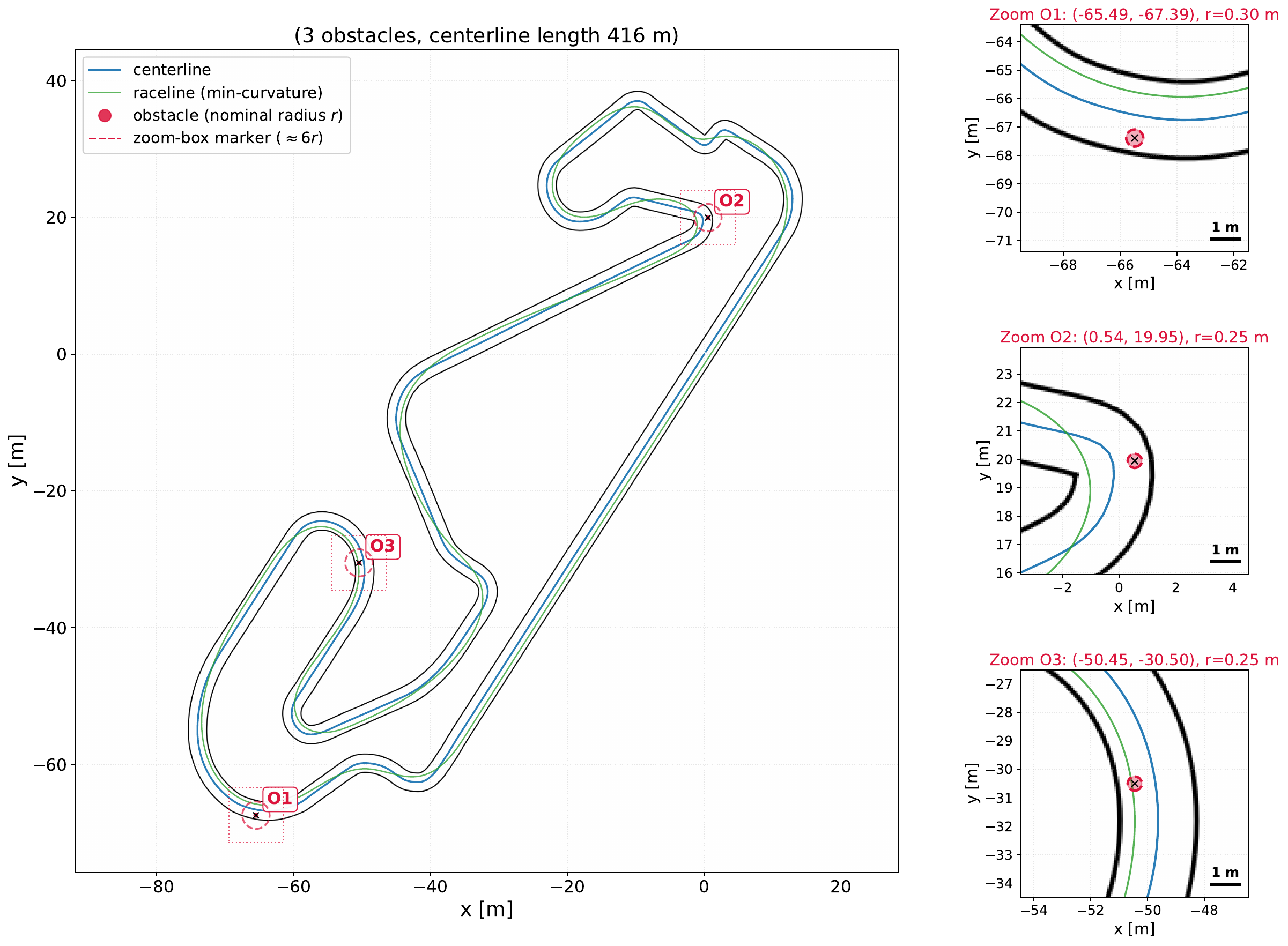}
  \caption{Catalunya (3 obstacles).}
  \label{fig:f1tenth_catalunya}
\end{subfigure}
\caption{F1TENTH track layouts. Blue: centerline; green: minimum-curvature
raceline; black: track walls. Red circles mark the safety-barrier
obstacles (radius $r$ in legend); zoom-in panels show each obstacle in
the context of the surrounding track walls. The safety filter constrains
only these super-ellipse obstacles; track-wall avoidance is encoded
implicitly through the demonstration distribution.}
\label{fig:f1tenth_obstacles}
\end{figure}

\paragraph{Environment.}
F1TENTH autonomous racing simulator~\cite{kelly2020f1tenth} with a
single-track bicycle dynamics model integrated via RK4. We evaluate on
two tracks: \emph{Budapest} (centerline length 402\,m, 1 obstacle) and
\emph{Catalunya} (centerline length 416\,m, 3 obstacles). Track
geometries and obstacle placements are shown in
Figure~\ref{fig:f1tenth_obstacles}. Evaluation over 100 rollouts per
track; the start and goal positions are perturbed by Gaussian noise with standard deviation $0.02$\,m.

\paragraph{Planning horizon and sampling steps.}
$H\!=\!696$ (Budapest), $H\!=\!724$ (Catalunya); $K\!=\!512$ for all
trajectory-level baselines.

\paragraph{Barrier and HOCBF parameters.}
Eq.~\eqref{eq:unified_barrier} with $n_i\!=\!2$ (circular, $r_{x,i}\!=\!r_{y,i}$)
and $\kappa\!=\!0.01$. The robust margin is shared between the safety
filter and the evaluation metric. For Safe-SFP variants, HOCBF gains are
$(\alpha_0, \alpha_1) = (100, 20)$ derived from $\alpha_{\text{cbf}} = 10$,
with trajectory step $\Delta t = 0.10$\,s. At racing speed, the drift term $L_f^2 h_i$ dominates the exponential CBF condition, so the filter omits this term for safety.

\paragraph{Evaluation mode.}
F1TENTH is evaluated in open loop, and a rollout reaches the goal when its final position lies within $1.0$\,m of the goal without a wall collision.
On Budapest, the plans of \ssf\ pass $6$--$12$\,cm outside the obstacle, but the car deviates from the plan by about $17$\,cm, which accounts for the remaining violations.

\subsection{Hopper}\label{app:hopper}

\paragraph{Environment.}
D4RL \texttt{hopper-medium-expert-v2}~\cite{fu2020d4rl}, a single-leg
locomotion task with MuJoCo physics. The 11-dimensional observation
excludes the global $x$-position for translational invariance:
\[
  o = [\underbrace{z,\ \alpha_y,\ q_{\text{thigh}},\ q_{\text{leg}},\ q_{\text{foot}}}_{\text{5-D position}},\;
       \underbrace{\dot{x},\ \dot{z},\ \dot{\alpha}_y,\ \dot{q}_{\text{thigh}},\ \dot{q}_{\text{leg}},\ \dot{q}_{\text{foot}}}_{\text{6-D velocity}}]
   \in \mathbb{R}^{11}.
\]
The action $a\in\mathbb{R}^3$ consists of thigh/leg/foot torques clipped to
$[-1,+1]^3$. Episode length $H_{\text{ep}}\!=\!1000$ control steps;
MuJoCo timestep $0.002$\,s with $\texttt{frame\_skip}\!=\!4$ (control step $0.008$\,s).

\begin{figure}[h]
\centering
\includegraphics[width=0.33\linewidth]{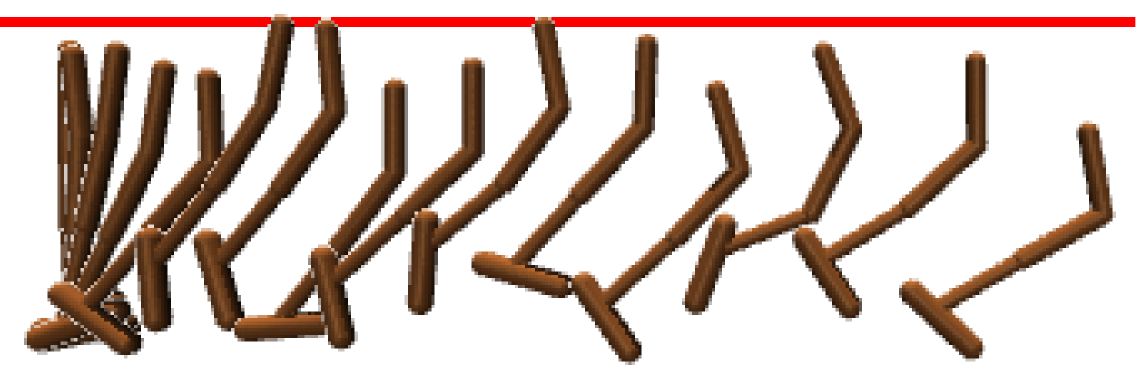}
\caption{Hopper rollout under \ssf with the velocity-augmented ceiling barrier (Eq.~\ref{eq:hopper_barrier}).
Red horizontal line: ceiling $z_{\max} = 1.5$\,m. The torso remains beneath the ceiling throughout the hopping gait.}
\label{fig:hopper_strip}
\end{figure}

\paragraph{Safety constraint.}
We enforce a velocity-augmented ceiling barrier
\begin{equation}\label{eq:hopper_barrier}
    \tilde{h}(\mathbf{x}) = z_{\max} - z - \beta\,\dot{z} - \delta,
    \qquad z_{\max} = 1.5\,\mathrm{m},\ \beta = 0.08,\ \delta = 0.01,
\end{equation}
together with a linear class-$\mathcal{K}$ gain $k = 0.15$ per control step ($18.75\,\mathrm{s}^{-1}$). The \emph{Viol.}\ column counts ceiling violations ($z\!>\!1.5$). A representative \ssf rollout under this barrier
is shown in Figure~\ref{fig:hopper_strip}.

\paragraph{Reduction from relative-degree-2 HOCBF to relative-degree-1 CBF.}
Under control-affine dynamics $\dot{\mathbf{p}}\!=\!\mathbf{v}$, $\dot{\mathbf{v}}\!=\!\mathbf{u}$, a position-only barrier $h_p(\mathbf{p})$ has relative degree 2: control $\mathbf{u}$ enters only at $\ddot{h}_p$, and the standard HOCBF chain requires two class-$\mathcal{K}$ conditions.
The velocity-augmented barrier~\eqref{eq:hopper_barrier} carries $\dot{z}$ in its definition, so its first time derivative already involves control:
\begin{align*}
    \dot{\tilde{h}} = -\dot{z} - \beta\,\ddot{z} = -\dot{z} - \beta\,u_{z},
\end{align*}
where $u_z$ is the component of $\mathbf{u}$ acting on vertical
acceleration. Hence $\tilde{h}$ has \emph{relative degree~1}, and the
two-level HOCBF chain collapses to a single class-$\mathcal{K}$ condition:
\begin{equation}\label{eq:hopper_cbf}
  \dot{\tilde{h}}(\mathbf{x}) + k\,\tilde{h}(\mathbf{x}) \;\ge\; 0,
  \qquad k = 18.75\,\mathrm{s}^{-1},
\end{equation}
enforced at every emission step. This is the relative-degree-1 case of the HOCBF framework.
Because the reduced constraint~\eqref{eq:hopper_cbf} is affine in the control input (the MuJoCo dynamics are linearized via a finite-difference Jacobian), the per-step problem is a \emph{quadratic program} (QP)---a quadratic tracking objective with one affine HOCBF constraint and box input bounds---rather than the QCQP used for the relative-degree-2 navigation tasks. We solve it with SLSQP.

\paragraph{Score metric.}
D4RL normalized return,
$\text{Score} = 100\!\times\!(R - R_{\text{random}})/(R_{\text{expert}} - R_{\text{random}})$,
averaged over 100 rollouts.

\paragraph{Timing.}
Hopper highlights the structural advantage of streaming.
As a receding-horizon task with no terminal goal, it requires a fresh control output at every step.
\ssf produces each action with a single streaming sample, so its per-action cost ($t_{\text{Opt}}$ for the filter, averaged over all steps; $t_{\text{total}}$ for the NN forward plus filtering) stays at the millisecond scale.
Trajectory-level baselines, by contrast, must generate---and safety-filter---a full $H$-step plan to emit even one action, inflating the per-action cost by orders of magnitude (e.g., SafeDiffuser's $1012.4$\,ms vs.\ \ssf's $7.30$\,ms).

\subsection{Warehouse Navigation}\label{app:gazebo}

\begin{figure}[h]
\centering
\includegraphics[width=0.75\textwidth]{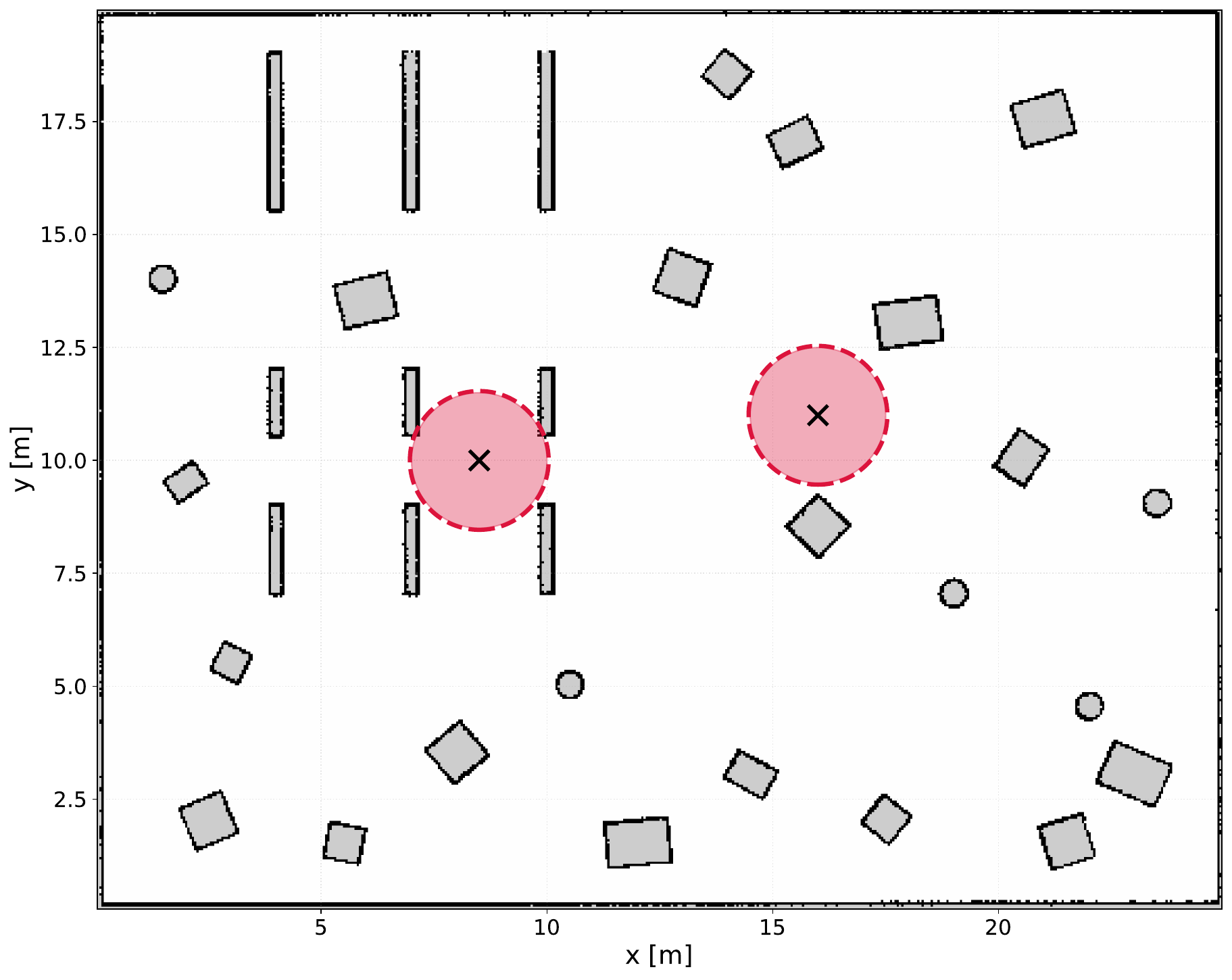}
\caption{Warehouse environment in Gazebo (top-down view). Gray polygons:
physical obstacles (shelves, pallets, crates) defined in the Gazebo SDF
world. Red dashed circles: abstract safety-filter barriers enforced via Eq.~\eqref{eq:unified_barrier} at $(8.5, 10.0)$ and $(16.0, 11.0)$ with radius $r=1.5$\,m. The two safety barriers are intentionally inflated relative to the physical shelf geometry to provide a margin and to match the line-density heatmap of training trajectories in the high-traffic central corridor.}
\label{fig:warehouse_obstacles}
\end{figure}

\begin{figure}[h]
\centering
\includegraphics[width=0.6\columnwidth]{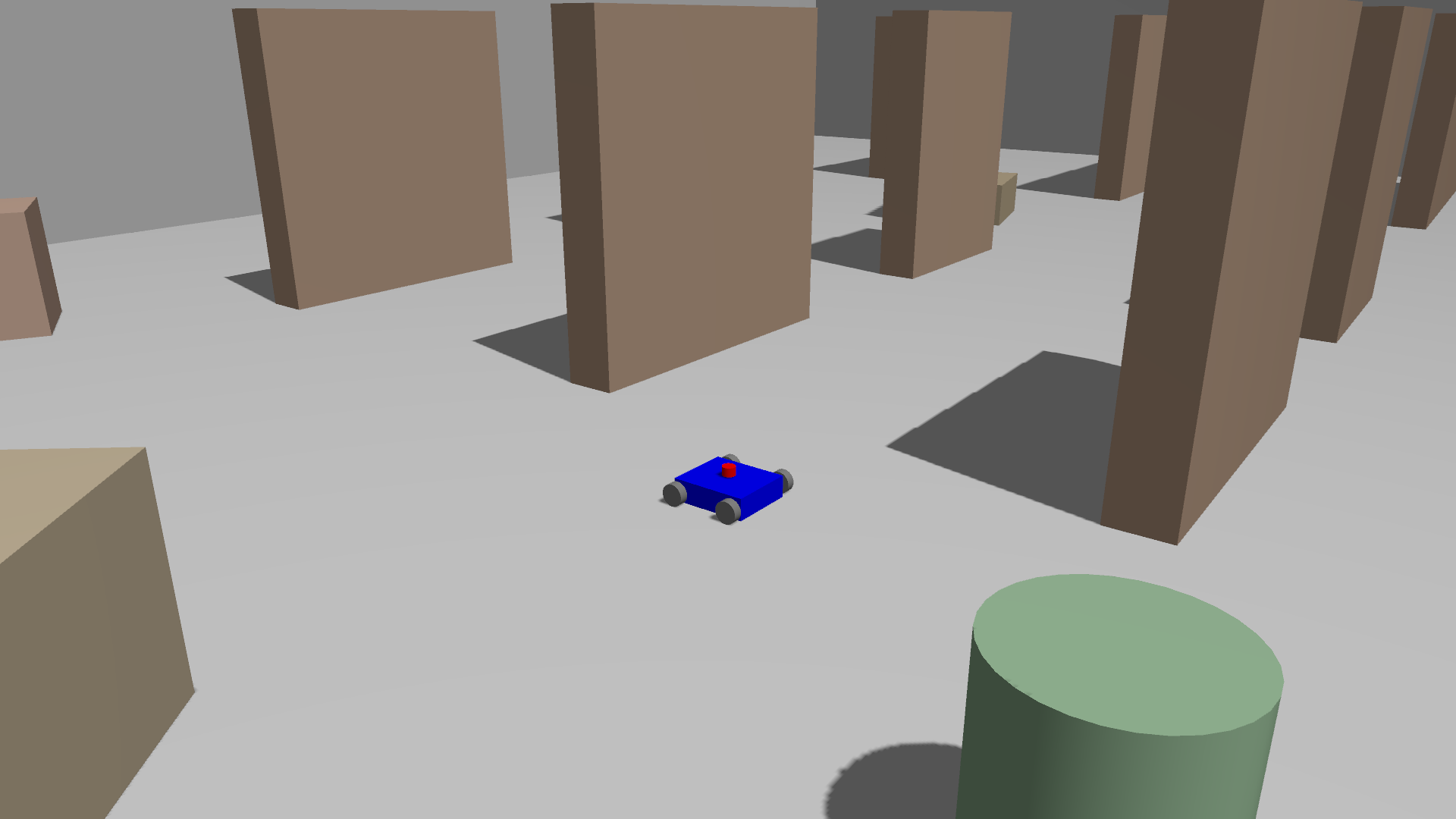}
\caption{Gazebo Fortress 3D view: the mecanum robot among warehouse obstacles,
simulated with a physics engine under the ROS2/Nav2 stack.}
\label{fig:gazebo_3d}
\end{figure}

\paragraph{Environment.}
ROS2-based Gazebo Fortress simulation of a $25\!\times\!20$\,m warehouse (free space $x\!\in\![1,24]$, $y\!\in\![1,19]$), populated with shelving, pallets, and crates as visualized in Figure~\ref{fig:warehouse_obstacles}.
The robot is a custom mecanum platform ($0.35\!\times\!0.30\!\times\!0.10$\,m chassis, $5$\,kg mass, $0.22$\,m Nav2 inflation radius). Physics step $4\!\times\!10^{-3}$\,s with real-time factor $1$.
The setup runs the full ROS2/Nav2 stack with rigid-body physics (Figure~\ref{fig:gazebo_3d}).

\paragraph{Demonstration data.}
$10{,}000$ trajectories collected by running Nav2~\cite{macenski2020marathon2}
on the warehouse with uniformly sampled random start and goal poses
(min goal distance $2.0$\,m), at a $20$\,Hz control rate.

\paragraph{Static obstacle layout.}
The safety filter treats two abstract circular obstacles in the
high-traffic corridor, $\mathcal{O} = \{((8.5, 10.0), r=1.5),\;((16.0, 11.0), r=1.5)\}$
with $n\!=\!2$ (see Figure~\ref{fig:warehouse_obstacles}).

\paragraph{Barrier and HOCBF parameters.}
Eq.~\eqref{eq:unified_barrier} with $n\!=\!2$ and $\kappa\!=\!0.05$; evaluation uses $\kappa\!=\!0$. For
Safe-SFP variants, HOCBF gains $(k_p, k_v) = (0.5, 0.3)$. Position-shield
QP variants (\texttt{safe\_diffuser}, \texttt{safe\_fm},
\texttt{safe\_flow\_matcher}) apply a one-step linearized projection to
each generated waypoint with endpoint re-pinning.

\paragraph{Controllers.}
All learned methods (Diffuser, FM, FlowMatcher, their safe variants, StreamingFlow, and \ssf) are executed through the same direct PD controller on \texttt{/cmd\_vel} at $20$\,Hz (gains $(K_p, K_v) = (2.0, 0.2)$, linear-velocity clamp $0.50$\,m/s).
The classical baselines (A*, RRT*) emit geometric paths without a velocity reference and are therefore tracked by the Nav2 \texttt{FollowPath} controller (Regulated Pure Pursuit, goal tolerance $0.25$\,m, desired linear velocity $0.35$\,m/s). For the streaming methods (StreamingFlow, \ssf), the PD loop additionally consumes the per-step velocity reference, and the HOCBF-QCQP filter (Eq.~\ref{eq:hocbf_qcqp}) operates inside this tick.

\paragraph{Benchmark protocol.}
100 (start, goal) pairs sampled with seed 42 from the free-space bounds above, distance range $[5,15]$\,m.
Goal tolerance $0.25$\,m.
After the planned horizon ($25.6$\,s), the PD controller keeps pulling the robot toward the goal.
The single violation of \ssf\ (closed-loop) occurs in this phase.
Sm.\ is the mean acceleration of the commanded velocity.

\end{document}